\documentclass{article}

\PassOptionsToPackage{sorting=none, style = numeric}{biblatex}

\usepackage[final]{neurips_2022}

\usepackage{microtype}
\usepackage{graphicx}
\usepackage{enumitem}
\usepackage{subfigure}
\usepackage{booktabs} 

\usepackage[utf8]{inputenc} 
\usepackage[T1]{fontenc}    
\usepackage{hyperref}       
\usepackage{url}            
\usepackage{booktabs}       
\usepackage{amsfonts}       
\usepackage{nicefrac}       
\usepackage{microtype}      
\usepackage{xcolor}         
\usepackage{amsmath}
\usepackage{hyperref}

\usepackage{amsmath,amsfonts,bm}

\def\eqref#1{equation~\ref{#1}}

\def\1{\bm{1}}

\DeclareMathAlphabet{\mathsfit}{\encodingdefault}{\sfdefault}{m}{sl}
\SetMathAlphabet{\mathsfit}{bold}{\encodingdefault}{\sfdefault}{bx}{n}

\usepackage{amssymb}
\usepackage{wrapfig}
\usepackage[utf8]{inputenc}
\usepackage[english]{babel}
\usepackage{amsthm}
\usepackage{adjustbox}
\usepackage{siunitx,pgf}

\usepackage[ruled,linesnumbered]{algorithm2e}
\usepackage{algorithm2e}

\usepackage{soul}
\usepackage{xr}
\usepackage{colortbl}
\usepackage{csquotes}
\usepackage{xcolor}

\usepackage{multirow}

\usepackage{caption}
\usepackage{multicol}

\makeatletter
\newcommand*{\addFileDependency}[1]{
  \typeout{(#1)}
  \@addtofilelist{#1}
  \IfFileExists{#1}{}{\typeout{No file #1.}}
}
\makeatother

\newif\ifdebug
\debugtrue
\ifdebug

\else

\fi

\usepackage{amssymb}
\usepackage{enumitem}
\usepackage{amsthm}
\usepackage{ulem}

\title{Learning NP-Hard Multi-Agent Assignment Planning using GNN: Inference on a Random Graph and Provable Auction-Fitted Q-learning}

\author{%
 Hyunwook Kang\thanks{First author} \\
  Department of Computer Science\\
  Texas A\&M University\\
  \texttt{hwkang@tamu.edu} \\
  \And
  Taehwan Kwon \\
  Kakao Brain \\
  isaac.kwon@kakaobrain.com  \\
  \AND
  Jinkyoo Park\thanks{correspondence to: Jinkyoo Park (jinkyoo.park@kaist.ac.kr), James R. Morrison (morri1j@cmich.edu)} \\
  Industrial \& Systems Engineering
  \\
  KAIST \\
  jinkyoo.park@kaist.ac.kr \\
  \And
  James R. Morrison$^{\dagger}$ \\
  Electrical Engineering \\
  Central Michigan University \\
  morri1j@cmich.edu \\
}

\begin{document}

\maketitle

\begin{abstract}
This paper explores the possibility  of  near-optimally  solving  multi-agent, multi-task NP-hard planning problems with time-dependent rewards using a learning-based algorithm. In particular, we consider a class of robot/machine scheduling problems called the multi-robot reward collection problem (MRRC). Such MRRC problems well model ride-sharing, pickup-and-delivery, and a variety of related problems. In representing the MRRC problem as a sequential decision-making problem, we observe that each state can be represented as an extension of probabilistic graphical models (PGMs), which we refer to as random PGMs. We then develop a mean-field inference method for random PGMs. We then propose (1) an order-transferable Q-function estimator and (2) an order-transferability-enabled auction to select a joint assignment in polynomial-time. These result in a reinforcement learning framework with at least $1-1/e$ optimality. Experimental results on solving MRRC problems highlight the near-optimality and transferability of the proposed methods. We also consider identical parallel machine scheduling problems (IPMS) and minimax multiple traveling salesman problems (minimax-mTSP).
\end{abstract}

\section{Introduction}
\textbf{Motivation} Consider a set of identical robots seeking to serve a set of spatially distributed tasks. Each task is given an initial age (which then increases linearly in time). Greater rewards are given to younger tasks when service is complete according to a predetermined reward rule. Such problems prevail in operations research, e.g., dispatching drivers to transport customers or scheduling machines in a factory.  Solving such highly structured NP-hard problems with the constraint of `no possibility of two robots assigned to one task at once' using mathematical optimization schemes is infeasible or ineffective due to the expensive computational cost, especially when the problem size is large. Applying decentralized approach using multi-agent modeling framework (\cite{long2020evolutionary, rashid2018qmix, sunehag2017valuedecomposition}) is a possible way to solve a large scale problems. However, due to the impossibility of inducing consensus among agents in achieving the global objective without effective communication (\cite{fischer1985impossibility}), such decentralized approaches are rarely used in industries (e.g., factories). Thus, this study focuses on centralized methods for solving MRRC.

\noindent
\textbf{Research Questions.} Many of such NP-hard scheduling problems have time-dependent rewards. To the best of our knowledge, these problems have not yet been addressed by non-decentralized learning-based methods. Even if one can, it must also be able to simultaneously address a fundamental challenge: the number of possible robot-task pairs to be considered increases exponentially. 
For instance, scheduling 8 robots and 50 tasks involves $10^{13}$ possible joint assignments  at each time-step. The main research question that the current study seeks to resolve is "how to design a computationally effective (i.e., scalability in terms of learning and decision making) a learning based centralized decision making scheme for solving a large-scale NP-hard scheduling problems?"

\noindent
\textbf{Proposed method and contributions.} The present paper explores the possibility of near-optimally solving multi-robot, multi-task NP-hard scheduling problems with time-dependent rewards using a learning-based algorithm. The study formulates multi-robot, multi-task NP-hard scheduling problems in a sequential decision-making framework and derives a joint scheduling policy with a theoretical performance bound under reasonable assumptions. The novelties of the current study are as follows:
\vspace{0.2cm}
\begin{itemize}[leftmargin=0.5cm]
\vspace{-0.1cm}
\setlength\itemsep{-0.1em}
    \item The study first observes that a state-joint assignment pair can be represented as a {\it random} PGM. After developing a theory of random PGM-based mean-field inference, we derive {\it random structure2vec}, a random PGM-based extension of structure2vec (\cite{Dai2016}).
    \item We estimate the $Q$-function $Q(s_k,a_k)$ using layers of {\it random structure2vec}, where $(s_k,a_k)$ is the state-joint action pair. Using an interpretation of a layer of structure2vec as a Weisfeiler-Lehman kernel (as in \cite{Dai2016}), we design the estimator to possess a property we call order-transferability. This property enables transferability in problem size.
    \item We propose a joint assignment rule called order-transferability-enabled auction policy (OTAP) to address exponential growth in joint assignment space. We propose auction-fitted $Q$-iteration (AFQI) by substitution of the $\operatorname{argmax}$ operation of fitted $Q$-iteration with OTAP to train the $Q$-function in a scalable manner. We prove that AFQI results in a policy with polynomial-time computation that achieves at least $1-1/e$ performance compared with the optimal policy.
\end{itemize}

\noindent
\textbf{Results and Impacts.}
Using simulation experiments, we show that the proposed policy typically achieves $97\%$ optimality for the multi-robot reward collection (MRRC) problem in a deterministic environment with linearly time-varying rewards. This performance is well extended to experiments with stochastic traveling times. To the best of our knowledge, this result is the first to learn a near-optimal NP-hard multi-robot/machine scheduling policy with time-dependent rewards.

\section{Related studies.} 

\textbf{Reinforcement Learning based Vehicle Routing Problems}. 
\citet{mazyavkina2020reinforcement} have categorized the RL approaches solving vehicle routing problems into two: (1) the improvement heuristics that learn an operator can iteratively improve the entire routing plans until there is no improvement is made \citep{wu2020learning, pmlr-v129-costa20a, NEURIPS2019_131f383b, Lu2020A, kim2021learning}, (2) the construction heuristics that learn a policy that sequentially make a single routing action given the partial solution (state) \citep{bello2016neural, nazari2018reinforcement, kool2018attention, khalil2017learning}, and (3) the hybrid approaches that mix these two approaches \citep{joshi2020learning, fu2021generalize, kool2021deep, pmlr-v119-ahn20a}. These RL approaches have mainly considered a single-agent routing problem. Although these methods solve CVRP with multiple vehicles, they solve this problem from a single-agent perspective. In addition, most of these approaches consider the static reward function setting. On the contrary, our study explicitly considers multi-vehicle interaction while considering the time-varying reward, which is a more realistic routing problem setting.

\textbf{Graph Inference Based Approach}.
\cite{Dai2017} showed that a graph neural network (GNN) called \textit{structure2vec} \cite{Dai2016} can construct a solution for the Traveling Salesman Problem (TSP). \textit{structure2vec} is a popular GNN derived from mean-field inference with a probabilistic graphical model (PGM). \cite{Dai2017} formulates the TSP as a Markov decision process (MDP) where a heuristically constructed PGM represents each state-next assignment pair. They employ \textit{structure2vec} derived from the heuristic PGM to infer the $Q$-function, which they use to select the next assignment. While their choice of PGM was heuristic, their approach achieved near-optimality and transferability of their trained single-robot scheduling algorithm to new single-robot scheduling problems with an unseen number of tasks.


\section{Multi-Robot Reward Collection Problem}
In the main text, we model a general multi robot/machine scheduling problem as a discrete-time, discrete-state (DTDS) sequential decision-making problem. In the DTDS model, time advances in fixed increments $\Delta$, i.e.,  $t_k=t_0+\Delta \times k$ where $t_k$ is the actual time after $k$ decision epochs have passed. For simplicity, we use $k$ as a time index representing the $k$th decision epoch. In this framework, $s_k$ denotes a state, and action $a_k$ denotes a joint assignment of robots/machines to unfinished tasks at the $k$th epoch. The objective of the problem is to learn the optimum scheduling policy $\pi_{\theta}:s_k\rightarrow a_k$ that maximizes the reward collected or minimizes the total completion time (a.k.a. makespan minimization). Below is the formulation of MRRC. We additionally propose a continuous-time continuous-state (CTCS) problem to identical parallel machine scheduling problems (IPMS) in Appendix \ref{IPMSappendix} and minimax multiple traveling salesman problems (minimax-mTSP) in Appendix \ref{minmaxmtsp}.

\subsection{State}
The state $s_k$ at epoch $k$ is represented as $(g_k, \mathcal{D}_k)$ where a graph $g_k=((\mathcal{R}, \mathcal{T}_k),(\mathcal{E}_{k}^{\mathcal{TT}}, \mathcal{E}_{k}^{\mathcal{RT}}) )$ and associated feature set $\mathcal{D}_k=(\mathcal{D}^{\mathcal{R}}_k,\mathcal{D}^{\mathcal{T}}_k,\mathcal{D}^{\mathcal{TT}}_k,\mathcal{D}^{\mathcal{RT}}_k)$. The elements of graph $g_k$ are defined as (See Figure 1):
\begin{itemize}[leftmargin=0.5cm]
\vspace{-0.15cm}
\setlength\itemsep{-0.1em}
    \item $\mathcal{R}=\{1,...,M\}$ is the index set of all robots. The index $i$ and $j$ will be used to specifically denote robots.
    \item $\mathcal{T}_k=\{1,...,N\}$ is the index set of all remaining unserved tasks at decision epoch $k$. The index $p$ and $q$ will be used to specifically denote tasks.
    \item $\mathcal{E}_{k}^{\mathcal{TT}}=\{\epsilon_{pq}^{\mathcal{TT}}|p\in\mathcal{T}_k,q\in\mathcal{T}_k\}$ is the set of all directed edges from a task in $\mathcal{T}_k$ to any task in $\mathcal{T}_k$. We consider each edge as a random variable. The task-to-task edge $\epsilon_{pq}^{\mathcal{TT}}=1$ indicates the event that a robot that has just completed task $p$ subsequently completes task $q$. We denote the probability  $p(\epsilon_{pq}^{\mathcal{TT}}=1)\in[0,1]$ the presence probability of the edge $\epsilon_{pq}^{\mathcal{TT}}$.
    \item $\mathcal{E}_{k}^{\mathcal{RT}}=\{\epsilon_{ip}^{\mathcal{RT}}|i\in\mathcal{R},p\in\mathcal{T}_k\}$ is the set of all directed edges from a robot in $\mathcal{R}$  to a task in $\mathcal{T}_k$. We say the robot-to-task edge $\epsilon_{ip}^{\mathcal{RT}}=1$ indicates the event that robot $i$ is assigned to the task $p$. This edge is defined deterministically depending on the joint assignment action. If robot $i$ is assigned to task $p$, then $p(\epsilon_{ip}^{\mathcal{RT}})=1$, otherwise 0.
\end{itemize}

\begin{figure*}[t!]
\centering
\includegraphics[width=1\textwidth]{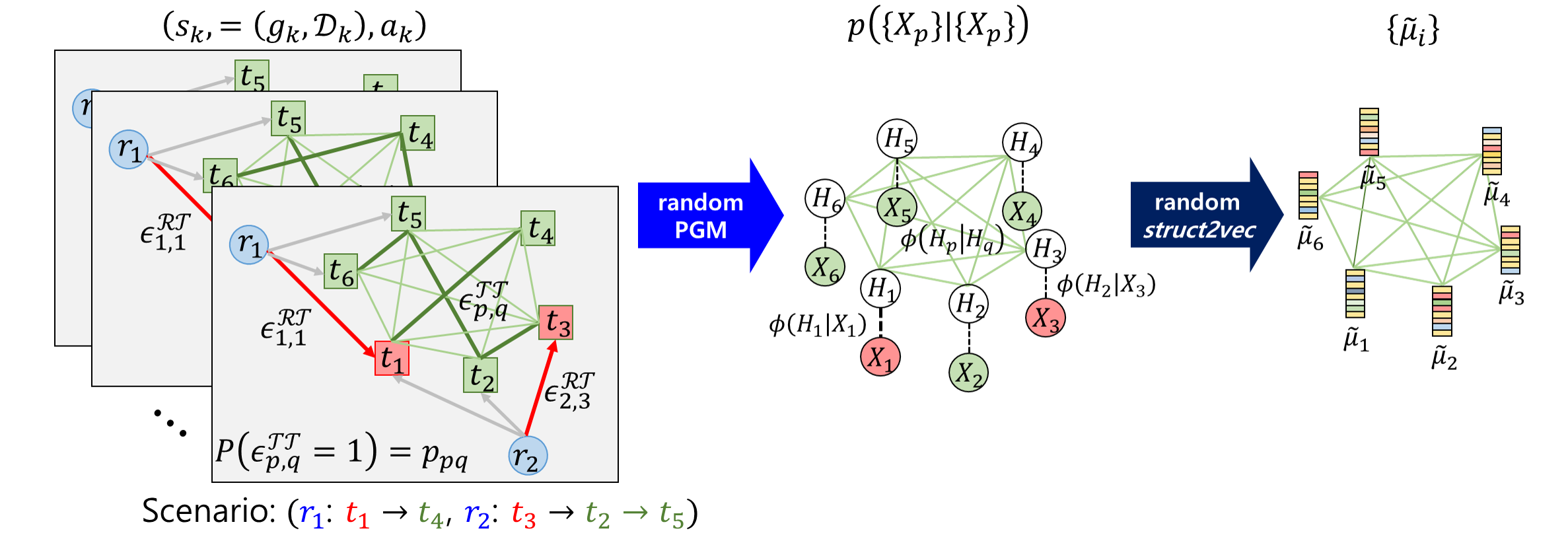}
\caption{Representation of the scheduling problem using a random PGM}
\label{PGM}  
\end{figure*}

\noindent
The element of feature set $\mathcal{D}_k$ associated with the graph $g_k$ is defined as:
\begin{itemize}[leftmargin=0.5cm]
\vspace{-0.1cm}
\setlength\itemsep{-0.1em}
    \item $\mathcal{D}^{\mathcal{R}}_k=\{d_i^{\mathcal{R}}|i\in \mathcal{R} \}$ is the set of node features for the robot nodes in $\mathcal{R}$ at epoch $k$. In MRRC, $d_i^{\mathcal{R}}$ is defined as the location of robot $i$ at epoch $k$ (epoch index $k$ is omitted).
    \item $\mathcal{D}^{\mathcal{T}}_k=\{d_p^{\mathcal{T}}|p\in \mathcal{T}_k \}$ 	is the set of node features for the task nodes in $\mathcal{T}_k$ at epoch $k$. In MRRC, $d_p^{\mathcal{T}}$ is defined as the age of task $p$ at epoch $k$ (epoch index $k$ is omitted).
    \item $\mathcal{D}^{\mathcal{TT}}_k=\{d_{pq}^{\mathcal{TT}}|p\in\mathcal{T}_k,q\in\mathcal{T}_k\}$ is the set of task-task edge features at epoch $k$. $d_{pq}^{\mathcal{TT}}$ denotes the duration for a robot that has just completed task $p$ to subsequently compete task $q$. 
    We call this duration \textit{task completion time.} In MRRC, a task completion time is given as a random variable (in practice, our method only requires a set of samples of random variable).
    \item $\mathcal{D}^{\mathcal{RT}}_k=\{d_{ip}^{\mathcal{RT}}|i\in\mathcal{R},p\in\mathcal{T}_k\}$ is the set of robot-task edge features at epoch $k$. $d_{ip}^{\mathcal{RT}}$ denotes the traveling time for robot $i$ to reach task $p$. 
\end{itemize}

\subsection{Action}
An action $a_k$, a joint assignment at epoch $k$, is defined as a maximal bipartite matching of the complete bipartite graph $(\mathcal{R}, \mathcal{T}_k,\mathcal{E}_{k}^{\mathcal{RT}})$ composed of the robot nodes $\mathcal{R}$, the remaining task nodes $\mathcal{T}_k$, and the fully connected edges between them $\mathcal{E}_{k}^{\mathcal{RT}}$. That is, given the current state $s_k=(g_k,\mathcal{D}_k)$, $a_k$ is a subset of  $\mathcal{E}_{k}^{\mathcal{RT}}$ satisfying (1) no two robots can be assigned to the same tasks, and (ii) a robot may only remain without assignment when the number of robots exceeds the number of remaining tasks. If $\epsilon_{ip}^{\mathcal{RT}}\in a_k$, it means that robot $i$ is assigned with task $p$ at epoch $k$. For example, Figure 1 shows the case where $a_k=(\epsilon_{1,1}^{\mathcal{RT}},\epsilon_{2,3}^{\mathcal{RT}})$. 
(Note, we equivalently may say this as $\epsilon_{1,1}^{\mathcal{RT}}=\epsilon_{2,3}^{\mathcal{RT}}=1$ and $0$ otherwise.)
In MRRC, all robots are allowed to change their assignments at each decision epoch. (In IPMS and mTSP, only free machines/salesmen are newly assigned.)

\subsection{State transition}
As the joint assignment $a_k$ is executed given the current state $s_k=(g_k,\mathcal{D}_k)$  , the next state $s_{k+1}=(g_{k+1},\mathcal{D}_{k+1})$ is determined with the updated graph $g_{k+1}$ and features $\mathcal{D}_{k+1}$.
\noindent

\textbf{Graph update.} When the decision epoch corresponds to the point when task $p$ is completed, the corresponding task node will be removed in the updated task nodes as
$\mathcal{T}_{k+1}=\mathcal{T}_{k}/\{p\}$, and the task-task edges and robot-task edges, $\mathcal{E}_{k+1}^{\mathcal{TT}}$ and $\mathcal{E}_{k+1}^{\mathcal{RT}}$, will be accordingly updated. 

\noindent
\textbf{Feature update.} At decision epoch $k+1$,  $\mathcal{D}_{k+1}=(\mathcal{D}^{\mathcal{R}}_{k+1},\mathcal{D}^{\mathcal{T}}_{k+1},\mathcal{D}^{\mathcal{TT}}_{k+1},\mathcal{D}^{\mathcal{RT}}_{k+1})$ is determined.  In MRRC, task locations $\mathcal{D}^{\mathcal{R}}_{k+1}=\{d_i^{\mathcal{R}}|i\in \mathcal{R}\}$ and task ages $\mathcal{D}^{\mathcal{T}}_{k+1}=\{d_p^{\mathcal{T}}|p\in \mathcal{T}_{k+1} \}$ are updated.
The robot-task edge features $\mathcal{D}^{\mathcal{RT}}_{k+1}$ will be updated according to $\mathcal{D}^{\mathcal{R}}_{k+1}$ as well. 

\subsection{Reward and objective}
At time $0$, each task is given an initial age which increases linearly in time. A reward $r_k=r\left(d_p^{\mathcal{T}}\right)$ is given when a task $p\in \mathcal{T}_k$, whose age is $d_p^{\mathcal{T}}$, is served at epoch $k$. We consider linear and nonlinear reward functions $r$ for MRRC. The objective is to learn a stationary policy $\pi$, a function that maps current state $s$ into current action $a$, to maximize expected total collected rewards $Q^{\pi}(s, a) =:E_{P, \pi}[\sum_{k=0}^\infty R(s_{t_k}, a_{t_k}, s_{t_{k+1}}) | s_{t_{0}}=s, a_{t_0}=a]  $.

\section{Random graph embedding: \textit{RandStructure2Vec}} 
We observe that when task completion time is not deterministic, a scheduling problem can be represented as an extension of probabilistic graphical models (PGMs), which we refer to as random PGMs. This section proposes a mean-field inference method, \textit{random struct2vec}, for random PGMs to estimate the state-action value $Q(s_k, a_k)$ in solving MRRC.

\subsection{Random PGM for representing a state of MRRC}
Given random variables $\mathcal{X} = \{X_p\}$, suppose that we can factor the joint distribution $p\left(\mathcal{X}\right)$ as  $p\left(\mathcal{X}\right)$$=$$\frac{1}{Z} \prod_{i} \phi_{i}\left(\mathcal{D}_{i}\right)$ where $\phi_{i}(\mathcal{D}_{i})$ denotes a marginal distribution or conditional distribution associated with a set of random variables $\mathcal{D}_{i}$; $Z$ is a normalizing constant. Then $\{X_p\}$ is called a probabilistic graphical model (PGM). In a PGM, $\mathcal{D}_{i}$ is called a clique and $\phi_{i}(\mathcal{D}_{i})$ is called the clique potential for $\mathcal{D}_{i}$, and $\mathcal{D}_{i}$ is called the scope of $\phi_{i}$. We often write simply $\phi_{i}$, suppressing $\mathcal{D}_{i}$.

Starting from a state $s_k$ and an action $a_k$, one can conduct a random experiment of ``sequential decision making using policy $\pi$''. In this random experiment, we can denote the events `How robots serve all remaining tasks in which sequence' as \textit{scenarios}. For example, suppose that at time-step $k$ we are given robots $\{r_1, r_2\}$, tasks $\{t_1, t_2, t_3, t_4, t_5, t_6\}$ and we follow the policy $\pi$ onward. One possible scenario is that robot $r_1$ serves tasks $\{t_1\rightarrow t_4\}$ and robot $r_2$ serves tasks $\{t_3\rightarrow t_2\rightarrow t_5\rightarrow t_6\}$ (see Figure 1). Note that the time when $t_5$ is served depends on the time when $t_2$ is served (state transition is Markovian); thus the reward from $t_5$ depends on the reward from $t_2$. As shown in Figure 1, $\{\{t_1\rightarrow t_4\},\{t_3\rightarrow t_2\rightarrow t_5\rightarrow t_6\}\}$ can be represented as a single instance of a Bayesian Network. Since scenario realization is random, we can construct the distribution over such scenarios using 'random' Bayesian Network with random node $X_k=(s_k, a_k)$ and clique potential $\phi$. For details, see section \ref{BNdetail}.

\subsection{Mean-field inference with random PGM} \label{str2vec}
Let $\mathcal{X}$ = $\{X_p\}$ be the set of all random variables in the inference problem. Let $\mathcal{G}_\mathcal{X}$ be the 
set of all possible PGMs on $\mathcal{X}$. Let $\mathcal{P}:\mathcal{G}_\mathcal{X} \mapsto [0,1]$ be a probability measure on $\mathcal{G}_\mathcal{X}$. Define a random PGM on $\mathcal{X}$ as $\{\mathcal{G}_\mathcal{X}, \mathcal{P}\}$. Note that the inference of $\{\mathcal{G}_\mathcal{X}, \mathcal{P}\}$ will be difficult; $|\mathcal{G}_\mathcal{X}|$ is too large for inferring $\mathcal{P}$ even using Monte-Carlo sampling approach. To avoid this difficulty, we use the approximated inference using {\it semi-cliques}. Suppose that we are given the set of all possible cliques on $\mathcal{X}$ as $\mathfrak{C}_{\mathcal{X}}$. As a PGM will be realized according to $\mathcal{P}$, only a few of the possible cliques in  $\mathfrak{C}_{\mathcal{X}}$ will be actually realized as an element of the PGM and become real cliques. We call such potential clique elements of $ \mathfrak{C}_{\mathcal{X}}$ as  {\it semi-cliques}. Note that if we were given $\mathcal{P}$, we could calculate the presence probability $p_m$ of the semi-clique $\mathcal{D}_{m}$ as $p_m = \sum_{G\in\mathcal{G}_\mathcal{X}}\mathcal{P}(G)\mathbf{1}_{\mathcal{D}_{m}\in G}$, where $\mathbf{1}$ denotes the indicator function. 

\vspace{0.3cm}
\noindent
\textbf{Mean-field inference with random PGM.} We start from a specific inference problem and state the main theorem in more general way. Consider a random PGM on  $\mathcal{X}=(\{H_i\}, \{X_j\})$ where $H_k$ is the latent variable corresponding to the observed variable $X_k$. Our goal is to infer $\{H_i\}$ given $\{X_j\}$ by finding $p(\{H_i\}|\{x_j\})$. In mean-field inference, we instead find a set of surrogate distributions $\{q^{\{x_j\}}(H_i)\}$ for which $\{H_i\}$ are independent. Here $q^{\{x_j\}}$ means that $q$ is a function of $\{x_j\}$). 

We next state \textit{\textbf{Theorem 1.}} in a very general manner. The statement is the same as that of mean-field inference with PGM (\cite{Koller}) except that ours has the presence probability terms $\{p_m\}$ of semi-cliques $\{\mathcal{D}_m\}$. The implication is that inference of presence probability of each semi-clique is enough to conduct mean-field inference, and the inference of $\{\mathcal{G}_\mathcal{X}, \mathcal{P}\}$ is not needed.

\vspace{0.3cm}
\noindent
\underline{\textit{\textbf{Theorem 1}. Random PGM based mean field inference.}} \textit{Suppose a random PGM on $\mathcal{X}=\{X_p\}$ is given, and the presence probability $\{p_{m}\}$ for all semi-cliques $\mathcal{D}_{m} \in \mathfrak{C}_{\mathcal{X}}$ are known. Then, the surrogate distribution $\{q_p (x_p)\}$ in mean-field inference is optimal only if $q_{p}\left(x_{p}\right)=\frac{1}{Z_{p}} \exp \left\{\sum_{m: X_{p} \in \mathcal{D}_{m}} p_{m} \mathbb{E}_{\left(\mathcal{D}_{m}-\left\{X_{p}\right\}\right) \sim q}\left[\ln \phi_{m}\left(\mathcal{D}_{m}, x_{p}\right)\right]\right\}$
where $Z_p$ is a normalizer and $\phi_{m}$ clique potential for $\mathcal{D}_m$.
}

For the general background and proof, see Appendix \ref{Thm1pf}.

\vspace{0.3cm}
\noindent
\textbf{RandStructure2Vec.}  In \cite{Dai2016}, structure2vec was derived as a vector space embedding of mean-field inference with PGM. For detailed background on vector space embedding, see Appendix \ref{embedding}. From \textit{\textbf{Theorem 1}}, we derive \textit{random structure2vec} as a vector-space embedding of mean-field inference with random PGM. Suppose that once a PGM is realized the PGM has joint distribution proportional to some factorization $\prod_{p} \phi\left(H_{p}| I_{p}\right) \prod_{p, q} \phi\left(H_{p}| H_{q}\right)$ (as in \cite{Dai2016}). Under this assumption, we can write $\{q^{\{x_j\}}(H_i)\}$ as $\{q^{x_i}(H_i)\}$. In \cite{Dai2016}, they suggest that structure2vec  is essentially a fixed point iteration ${\tilde{\mu}}_{p}\leftarrow\sigma(W_{1} x_{p}+W_{2} \sum_{q \neq p}  {\tilde{\mu}}_{q})$ where $\tilde{\mu}_p$ is a latent vector for node $p$ and $x_p$ is input for node $p$. They show that, when we interpret $\tilde{\mu}_i$ as a vector space injective embedding expressed as $\tilde{\mu}_i=\int_{\mathcal{H}} \phi(h_i)q^{x_i}(h_i) dh_i$ for some $\phi$, structure2vec's fixed point iteration is the embedding of fixed point iteration of mean-field inference with the PGM. \textit{Lemma 1} states that we have a similar result for random PGM by including only the $\{p_{qp}\}$ information.

\vspace{0.3cm}
\noindent
\textit{\underline{\textbf{Lemma 1}. Structure2vec for random PGM.} \label{Lemma 1} Assume that the presence probabilities $\{p_{qp}\}$ for all pairwise semi-cliques $\mathcal{D}_{qp} \in \mathfrak{C}_{\mathcal{X}}$ are given. Then embedding the fixed point equation in \textit{Theorem 1} generates the fixed point equation $
     {\tilde{\mu}}_{p}\leftarrow \sigma\left(W_{1} x_{p}+W_{2} \sum_{q \neq p} p_{qp} {\tilde{\mu}}_{q}\right)
$. We refer to this fixed point iteration as \textit{random structure2vec}. (The proof of \textit{Lemma 1} can be found in Appendix \ref{Lem1pf}.)} 

\vspace{0.3cm}
\noindent
\textbf{Remarks.} Note that inference of $\{\mathcal{G}_{\mathcal{X}},\mathcal{P}\}$ is in general a difficult task. One implication of \textit{Theorem 1} is that \textit{we transformed a difficult inference task into a simple inference task}: inferring the presence probability of each semi-clique. (See Appendix \ref{pinference} for the algorithm that conducts this task.) In addition, \textit{Lemma 1} provides a theoretical justification to ignore the inter-dependencies among edge presences when embedding a random graph using GNN. When graph edges are not explicitly given or known to be random, the simplest heuristic one can use is to separately infer the presence probabilities of all edges and adjust the weights of GNN's message propagation. According to \textit{Lemma 1}, possible inter-dependencies among edges would not affect the quality of such heuristic inference.

\section{Solving MRRC with \textit{RandStructure2Vec}}
\begin{figure*}[h!]
\centering
\includegraphics[width=1.0\textwidth]{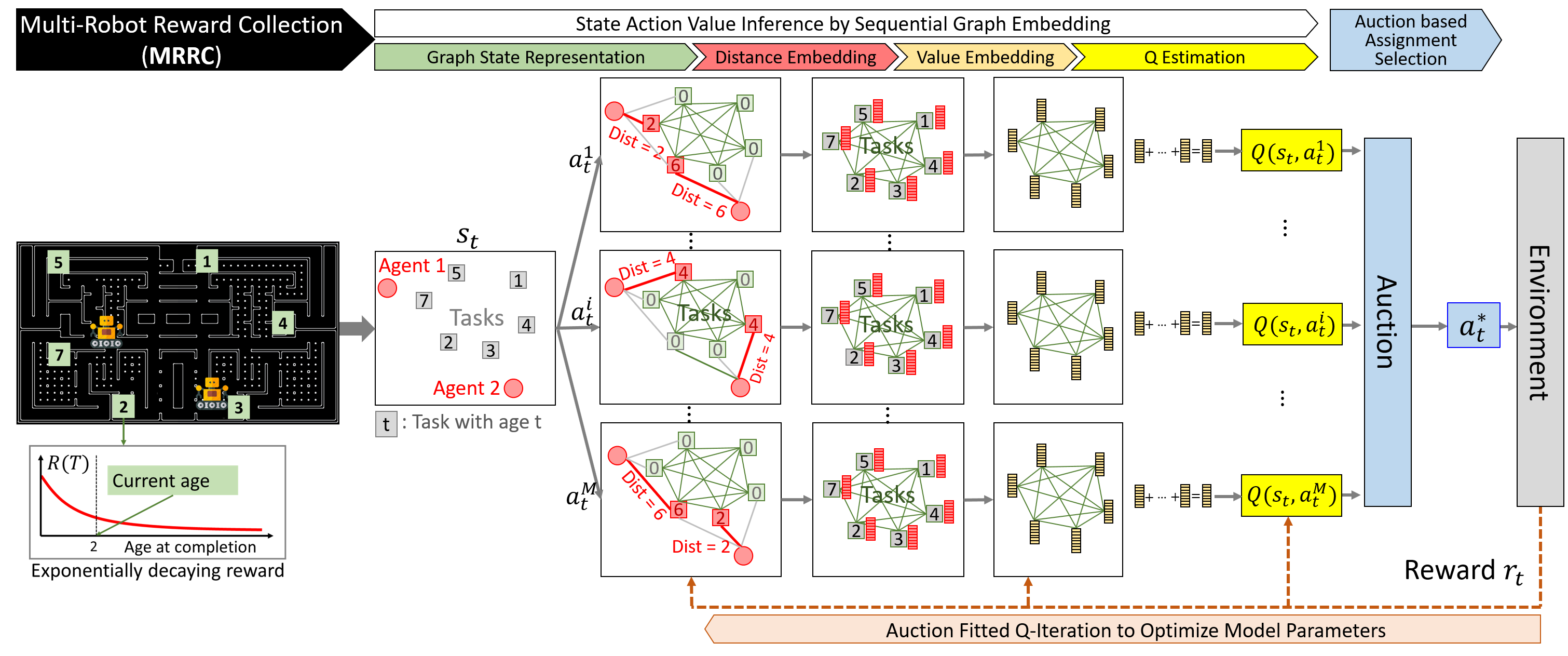}
\caption{State representation and main inference procedure}
\label{PGM}  
\end{figure*}

This section describes how the proposed method solves the MRRC problem with the newly developed \textit{random structure2vec} method. The total solution (i.e., the sequential assignments of robots/machines to tasks) is found by iteratively repeating a sequential decision making. This section specifically describes how to choose a joint assignment $a_k$ given the current state $s_k=(g_k, \mathcal{D}_k)$. The procedure for determining a joint assignment action is composed of (1) represent the state using a random Bayesian Network, (2) estimate the Q-value using random graph embedding, and (3) select a joint assignment. Figure 2 depicts the overall procedure. This section focuses on MRRC; the appendices will provide the formulation, solution procedure, and results for IPMS and mTSP problems as well.

\subsection{Representing a state using a random PGM}\label{BNdetail}
In this section, we describe how the MRRC problem state $s_k$ and one possible joint assignment $a_k$ can be represented as a Bayesian network. Assume $H_p$, the hidden random variable for task $p$, carries information about the benefit of serving task $p$. Given a scenario (see section 3.1), $H_p$ depends on $H_q$ only if $q$ is served after $p$ by the same robot, and $H_p$ depends on $X_p$, the observed features for task $p$. For an MRRC problem, $X_p$ can be either a task's assignment information $d_{ip}^{\mathcal{RT}}$ or a task's age $d_p^{\mathcal{T}}$. Based on these definitions, a Bayesian Network for $\{H_p\}$ and $\{X_p\}$ is constructed as $p(\{H_p\}|\{X_p\})=\prod_{p} \phi\left(H_{p}|X_{p}\right) \prod_{p, q} \phi\left(H_{p}| H_{q}\right)$. This Bayesian Network corresponds to one scenario; there are many possible scenarios that can be realized randomly. Since random PGM naturally models this characteristic, according to Lemma 1, we are justified to use \textit{random structure2vec} with edge (semi-clique) presence probabilities $\{p(\epsilon_{pq}^{\mathcal{TT}})\}$ from in section 2.1. 

\subsection{Estimating state-action value using Order trainability-enabled Q-function}\label{Qinference}
We illustrate how we can estimate a Q-function for MRRC by applying \textit{random structure2vec} to the random PGM that represents ($s_k, a_k$). \textit{Lemma 1} provides the fixed point equation $
     {\tilde{\mu}}_{p}=\sigma\left(W_{1} x_{p}+W_{2} \sum_{j \neq k} p_{qp} {\tilde{\mu}}_{q}\right)
$to compute the embeddings $\tilde{\mu}_{p}$ for node task $p$ in a random PGM. The embeddings $\tilde{\mu}_{p}$ for $p \in \mathcal{T}_k$ are computed in an iterative manner using \textit{random structure2vec} as: $
{\tilde{\mu}}_{p}^{(\tau+1)}=\sigma\left(W_{1} x_{p}+W_{2} \sum_{q \neq p} p_{qp} {\tilde{\mu}}_{q}^{(\tau)}\right)$. We propose a network architecture composed of two-step sequential \textit{random structure2vec}, which is composed of \textit{action embedding} and \textit{value embedding}.

\noindent
\textbf{Action Embedding.} The first \textit{random structure2vec} layer embeds the joint assignment into the task nodes $\mathcal{T}_k=\{1,...,n\}$. The action embedding $\tilde{\mu}^A_p$ for task node $p$ is defined as the fixed point for the equation $\tilde{\mu}^A_p=\sigma\left(W^A_{1} x_{p}^A+W^A_{2} \sum_{q \neq p} p_{qp} {\tilde{\mu}}^{A}_{q}\right)$ where $x_p^A=d_{ip}^{\mathcal{RT}}$ (distance from robot $i$ to task $p$) when task $p$ is assigned to robot $i$, $x^A_p=0$ when task $p$ is not assigned. The action node embeddings $\{\tilde{\mu}^A_p|p \in \mathcal{T}_k\}$, computed via iteration simultaneously  by employing \textit{random structure2vec}, provide sufficient information about the relative locations between robots and their assigned tasks.

\noindent
\textbf{Value Embedding.} The second \textit{random structure2vec} layer embeds the task ages into the task nodes. The value embedding $\tilde{\mu}^V_p$ for task node $p$ is defined as the fixed point for the equation $\tilde{\mu}^V_p=\sigma\left(W^V_{1} x_{p}^V+W^V_{2} \sum_{q \neq p} p_{qp} {\tilde{\mu}}^{V}_{q}\right)$ where $x_p^V=(\tilde{\mu}^A_p, d_{p}^{\mathcal{T}} )$ is the concatenation of the action embedding $\tilde{\mu}^A_p$ computed by the first layer and the task age $d_{p}^{\mathcal{T}}$ for node $p$. The resulting value node embeddings
$\{\tilde{\mu}^V_p|p \in \mathcal{T}_k\}$, computed in an iterative manner, provide sufficient information about how much value is likely in the local graph around each task by the specified joint assignment.

\noindent
\textbf{Computing $Q_{\theta}(s_{k},a_{k})$.} To derive $Q_{\theta}(s_{k}, a_{k})$, we aggregate the embedding vectors for all nodes by $\tilde{\mu}^V = \sum_p \tilde{\mu}^{V}_p$ to obtain one global vector $\tilde{\mu}^V$ to embed the value affinity of the global graph. We then use a neural network to map $\tilde{\mu}^V$ into $Q_\theta(s_{k}, a_{k})$. The overall pseudo-code for estimating steps is provided in Appendix \ref{randomsample}. 

It is essential for the proposed inference method can estimate the state action value $Q_{\theta}(s_{k},a_{k})$ for varying graph size $g_k$. Let us provide the intuition related to scuh problem-size transferability. For \textit{Action Embedding}, transferability is trivial; the inference problem is a scale-free task {\it locally around each node}. For \textit{Value Embedding}, consider the ratio of robots to tasks. The overall value affinity embedding will be underestimated if this ratio in the training environment is smaller than this ratio in the testing environment; overestimated overall otherwise. The intuition is that this over/under-estimation does not matter in Q-function based policies as discussed in \cite{VanHasselt2015} as long as the \textit{order} of Q-function value among actions are the same. That is, as long as the best assignments chosen are the same, i.e., $\arg\max_{a_{k}} Q(s_{k}, a_{k})$ = $\arg\max_{a_{k}} Q_{\theta}(s_{k}, a_{k})$, the magnitude of imprecision $|Q(s_{k}, a_{k}) -  Q_{\theta}(s_{k}, a_{k})|$ does not matter. We call this property of an estimator \textit{order-transferability} (with respect to the max operation).

\subsection{Selecting a joint assignment using OTAP}
With the previously introduced way to estimate $Q_{\theta}(s_{k},a_{k})$, we illustrate how to compute the joint assignment (action) $a_k^*$, a maximal bipartite matching in the bipartite graph $(\mathcal{R}, \mathcal{T}_k,\mathcal{E}_{k}^{\mathcal{RT}})$, given the state $s_k=(g_k,D_k)$. Specifically, we propose the order transferability-enabled auction policy (OTAP) that constructs a joint assignment $a_k$ through $N=\max \left(|\mathcal{R}|,\left|k\right|\right)$ iterations of \textit{Bidding} and \textit{Consensus} phases. This auction follows the spirit of Sequential Single Item (SSI) auctioning (\cite{koenig2006power}); each iteration adds one robot-task assignment to construct a full joint assignment.

\noindent
\textbf{Bidding-phase.} In the $n^{th}$ iteration of the bidding phase, given $\mathcal{M}_\theta^{(n-1)}$, the ordered set of $n-1$ robot-task edges in $\mathcal{E}_{k}^{R T}$ determined by the previous $n-1$ iterations,  all the unassigned robots bid their the most preferable task to conduct. Respecting robot-task assignments determined in previous $n-1$ iterations and ignoring other unassigned robots, each unassigned robot $i$ select the best task assignment $\epsilon_{il}^{\mathcal{RT}}$ that maximizes $Q_{\theta}^n(s_k,\mathcal{M}_\theta^{(n-1)} \cup \{ \epsilon_{ip}^{\mathcal{RT}} \})$ among all unassigned tasks $p\in \mathcal{T}_k$, where $Q_{\theta}^n$ is the $\theta$-parameterized network with superscript $n$ indicating that the state action value is estimated at $n^{th}$ iteration. Then, robot $i$ bids $\{\epsilon_{i \ell}^{\mathcal{RT}},Q_\theta^n(s_k, \mathcal{M}_\theta^{(n-1)}\cup \{ \epsilon_{i \ell}^{\mathcal{RT}} \} )\}$ to the auctioneer. This bidding occurs simultaneously by all the unassigned robots at the $n^{th}$ iteration. Since the number of ignored robots varies at each iteration, transferability of Q-function inference is crucial.



\noindent
\textbf{Consensus-phase.} In the consensus phase of  $n^{th}$ iteration, the centralized auctioneer finds the bid with the best bid value, say $\{\epsilon_{i^*p^*}^{\mathcal{RT}},Q_\theta^n(s_k, \mathcal{M}_\theta^{(n-1)}\cup \{ \epsilon_{i^* p^*}^{\mathcal{RT}} \} )\}$ (Here $i^*$ and $p^*$ denote the best robot task pair.) Denote $\epsilon_{i^*p^*}^{\mathcal{RT}} := m_\theta^{(n)}$. The auctioneer sets everyone's $\mathcal{M}_\theta^{(n)}$ as $\mathcal{M}_\theta^{(n)}=\mathcal{M}_\theta^{(n-1)} \cup m_\theta^{(n)}$ and initiate bidding phase for the remaining unassigned robots.

These two phases iterate until reaching $\mathcal{M}_\theta^{(N)} = \{m_\theta^{(1)}, \dots, m_\theta^{(N)}\}$. This $\mathcal{M}_\theta^{(N)}$ is chosen as the joint assignment $a_{k}^*$ of $N$-robots at time step $k$. That is, $\pi_{\theta}(s_{k}) = a_{k}^*$. The computational complexity for computing $\pi_{\theta}$ is $\mathrm{O}\left(\left|{R}\right|\left|{T_{}}\right|\right)$ and is only polynomial; see Appendix \ref{complexity}.

\subsection{Training Q-function using AFQI}
The fitted Q-iteration (FQI) finds $\theta$ that minimizes $E_{(s_k, a_k, r_k, s_{k+1})\sim D }$ $[Q_{\theta}\left(s_{k}, a_{k}\right)-[r\left(s_{k}, a_{k}\right)+\gamma \max_a Q_{\theta}\left(s_{k+1}, a\right)]]$ where $D$ denotes the distribution of training data. We propose a new reinforcement learning method, which we call \textit{Auction-fitted Q-iteration (AFQI)}, which replaces $\operatorname{max}_{a}$  $Q_\theta\left(s_{k}, a\right)$ used in the conventional FQI with OTAP. That is, writing OTAP as $\pi_{Q_\theta}$, AFQI finds $\theta$ that empirically minimizes $E_{(s_k, a_k, r_k, s_{k+1})\sim D }$ $\left[Q_{\theta}\left(s_{k}, a_{k}\right)-\left[r\left(s_{k}, a_{k}\right)+\gamma Q_{\theta}\left(s_{k+1}, \pi_{Q_{\theta}}\left(s_{k+1}\right)\right)\right]\right].$

In learning the parameters $\theta$ for $Q_{\theta}\left(s_{k}, a_{k}\right)$, we use the exploration strategy that perturbs the parameters $\theta$ randomly to actively explore the joint assignment space with OTAP. While this method was originally developed for policy-gradient based methods \cite{Plappert2017}, exploration in parameter space is useful in our auction-fitted Q-iteration since it generates a reasonable combination of assignments.

\section{Theoretical analysis}

We show the proposed AFQI obtains at least $1-1/e$ optimality and enables computation of the joint assignment in polynomial time. This result is achieved by the order-transferability of the proposed $Q$-function estimator and its use in selecting the joint assignment.

\subsection{Performance bound of OTAP}
Recall that $Q^n$ denotes the $n$-robot problem's true $Q$-function. In the same way as we defined $\mathcal{M}_\theta^{(N)}$ above, denote the joint assignment chosen by OTAP as $\{Q^{n}\}_{n=1}^{N}$  as $\mathcal{M}^{(N)}$ $=$ $\{m^{(1)}, \dots, m^{(N)}\}$.

\noindent
\textit{\underline{\textbf{Lemma 2.}} If the $Q$-function approximator has order transferability, then $\mathcal{M}^{(N)}$ = $\mathcal{M}_\theta^{(N)}$.
}

For any decision epoch $k$, let $\mathcal{M}$ denote a set of robot-task pairs (a subset of $\mathcal{E}_{k}^{\mathcal{RT}}$). For any robot-task pair $m \in \mathcal{E}_{k}^{\mathcal{RT}}$, define $\Delta(m\mid \mathcal{M}) := Q^{|\mathcal{M} \cup\{m\}|}(s_{k}, \mathcal{M} \cup\{m\})-Q^{|\mathcal{M}|}(s_{k}, \mathcal{M})$ as the the marginal value (under the true Q-functions) of adding robot-task pair $m \in \mathcal{E}_{k}^{\mathcal{RT}}$. Lemma 2 enables us to use the result discussed in \cite{nemhauser1978analysis} and achieve the result of Theorem 2.

\noindent
\textit{\underline{\textbf{Theorem 2.}} Suppose that the $Q$-function approximation with the parameter value $\theta$ exhibits order transferability. Denote $\mathcal{M}_{\theta}^{(N)}$ as the result of OTAP using $\left\{Q_{\theta}^{n}\right\}_{n=1}^{N}$ and let $\mathcal{M}^{*}$ $=$ $\operatorname{argmax}_{a_{k}}$ $Q^{|a_{k}|}    \left(s_{k}, a_{k}\right)$.
If $\Delta(m\mid \mathcal{M}) \ge 0, \forall \mathcal{M} \subset \mathcal{E}_{k}^{R T}, \forall m \in \mathcal{E}_{k}^{\mathcal{RT}}$, and the marginal value of adding one robot diminishes as the number of robots increases, i.e., $\Delta(m\mid \mathcal{M})$ 
$\le$ 
$\Delta(m \mid \mathcal{N}), \forall \mathcal{N} \subset \mathcal{M}\subset \mathcal{E}_{k}^{R T}$, $\forall m \in \mathcal{E}_{k}^{R T}$, 
then the result of OTAP is at least better than $1-1/e$ of an optimal assignment. That is, $Q_{\theta}^{N}(s_{k}, \mathcal{M}_{\theta}^{(N)}) $$\ge$$ Q^{|\mathcal{M}^{*}|}\left(s_{k}, \mathcal{M}^{*}\right)$$\left(1-1/e\right).$ }
See Appendix \ref{Lem2pf} and \ref{Thm2pf} for the proofs.

\subsection{Performance bound of AFQI}
AFQI seeks to find $\theta$ that minimizes $E_{(s_k, a_k, r_k, s_{k+1})\sim D }$ $[Q_{\theta}\left(s_{k}, a_{k}\right)-[r\left(s_{k}, a_{k}\right)+\gamma Q_{\theta}\left(s_{k+1}, \pi_{Q_{\theta}}\left(s_{k+1}\right)\right)]].$ 
Here, we use OTAP, denoted as $\pi_{Q_\theta}$, instead of   $\operatorname{max}_{a}$  $Q_\theta\left(s_{k}, a\right)$ which is used in general fitted-Q iteration. As we have seen in section 5.1, OTAP replaces the \textit{max} operation with the auction algorithm with a provable performance bound compared with the \textit{max} operation. \textit{Lemma 3} allows us to use this performance bound to obtain a performance assurance on AFQI compared with FQI. 
We only write an abbreviated version of the statement for brevity. The formal description of conditions and the proof is provided in Appendix \ref{lemma3}.

\noindent
\underline{\textit{\textbf{Lemma 3.} \cite{kang2021approximate}}} Suppose that a $1-1/r$ approximation algorithm is substituted for the $\max$ operation in FQI. Then, the corresponding new Fitted Q-iteration's performance is at least $1-1/r$ optimal.

\noindent
\underline{\textit{\textbf{Corollary 1.}}} AFQI achieves at least $1-1/e$ performance compared with the optimal policy.


\begin{table*}[h!]
\centering
\scriptsize
\caption{Performance test (50 trials of training for each case)}
\label{Performance-table}
\def\arraystretch{1.0}
\makebox[\textwidth][c]{
\begin{tabular}{cccccccccc}
\toprule
\multirow{2}{*}{\textbf{Reward}}        &\multirow{2}{*}{\textbf{Environment}} & \multirow{2}{*}{\textbf{Baseline}}   & \multicolumn{7}{c}{\textbf{Testing size : Robot (R) / Task (T)}}                \\
 & & & 2R/20T  & 3R/20T  & 3R/30T  & 5R/30T  & 5R/40T  & 8R/40T  & 8R/50T  \\ \toprule
\multirow{8}{*}{Linear} & \multirow{6}{*}{Deterministic} & \multirow{2}{*}{Optimal} & 98.31  & 97.50 & 97.80 & 95.35  & 96.99  & 96.11 & 96.85 \\
& &  &($\pm4.23$) &($\pm4.71$)& ($\pm5.14$)& ($\pm5.28$)& ($\pm5.42$)& ($\pm4.56$) & ($\pm3.40$) \\ 
\cline{3-10}
                              & & \multirow{2}{*}{Ekisi et al.}           & 99.86& 97.50& 118.33& 110.42& 105.14& 104.63 & 120.16 \\
                              & &    & ($\pm3.24$)& ($\pm2.65$)& ($\pm2.84$)& ($\pm2.97$)& ($\pm3.78$)& ($\pm2.50$) & ($\pm3.94$) \\ \cline{3-10} 
                              & & \multirow{2}{*}{SGA}   & 137.3 & 120.6 & 129.7 & 110.4 & 123.0 & 119.9 & 119.8 \\
                              & &  & ($\pm5.65$)&($\pm5.03$)& ($\pm5.54$)& ($\pm4.34$)& ($\pm4.97$)& ($\pm4.74$) & ($\pm5.84$) \\ 
                              \cline{2-10}
                              & \multirow{2}{*}{Stochastic} & \multirow{2}{*}{SGA}   & 130.9 & 115.7 & 122.8 & 115.6 & 122.3 & 113.3 & 115.9 \\
& &  & ($\pm4.02$)& ($\pm4.03$)& ($\pm5.21$)& ($\pm6.23$)& ($\pm4.94$)& ($\pm5.53$) & ($\pm4.08$) \\                               
                              \hline
\multirow{4}{*}{Nonlinear} & \multirow{2}{*}{Deterministic} & \multirow{2}{*}{SGA}   & 111.5 & 118.1 & 118.0 & 110.9 & 118.7 & 111.2 & 112.6 \\ 
 & & & ($\pm3.71$) &($\pm5.56$) & ($\pm5.09$) & ($\pm4.64$)& ($\pm5.23$) & ($\pm5.38$) & ($\pm5.07$) \\
\cline{2-10} 
                              & \multirow{2}{*}{Stochastic} & \multirow{2}{*}{SGA}   & 110.8 & 117.4 & 119.7 & 111.9 & 120.0 & 110.4 & 112.4 \\
    &&  & ($\pm5.17$) & ($\pm6.22$) & ($\pm4.48$) & ($\pm4.70$) & ($\pm6.38$) & ($\pm5.14$) & ($\pm5.30$) \\
    \bottomrule
\end{tabular}}
\end{table*}

\begin{table*}[h!]
\centering
\caption{Transferability test (linear \& deterministic env, standard dev. provided in the appendix)}
\label{transfer}
\scriptsize
\def\arraystretch{1.5}
\makebox[\textwidth][c]{
\begin{tabular}{cccccccc} \toprule
\textbf{Training size} & \multicolumn{7}{c}{\textbf{Testing size : Robot (R) / Task (T)}}              \\
\textbf{\#R/ \#T} & 2R/20T  & 3R/20T  & 3R/30T & 5R/30T  & 5R/40T  & 8R/40T  & 8R/50T  \\\bottomrule  
2R/20T   & \cellcolor{gray}{98.31 ($\pm4.23$)}  & 93.61($\pm4.98$)  & 97.31 ($\pm4.25$)  & 92.16 ($\pm3.49$)   & 92.83($\pm4.25$)  & 90.94($\pm3.98$)  & 93.44 ($\pm4.02$)  \\  
3R/20T   & 95.98($\pm4.75$)  & \cellcolor{gray}{97.50($\pm3.71)$}  & 96.11($\pm3.63$)   & 93.64($\pm4.54$)  & 91.75($\pm5.71$)  & 91.60($\pm5.03$)  & 92.77($\pm4.74)$  \\  
3R/30T   & 94.16($\pm4.97$)& 96.17($\pm4.22$)  & \cellcolor{gray}{97.80($\pm5.14)$}   & 94.79($\pm3.53$)  & 93.19($\pm3.78$)  & 93.14($\pm4.50$)  & 93.28($\pm3.99$)  \\  
5R/30T   & 97.83($\pm3.11$)  & 94.89($\pm4.43$)  & 96.43($\pm4.23$)   & \cellcolor{gray}{95.35$\pm5.28)$}  & 93.28($\pm4.18$)  & 92.63($\pm5.07$)  & 92.40($\pm4.10$)  \\  
5R/40T   & 97.39($\pm4.65$)  & 94.69($\pm4.01$)  & 95.22($\pm4.88$)   & 93.15($\pm5.09$)  & \cellcolor{gray}{96.99$\pm4.42)$}  & 94.96($\pm3.94$)  & 93.65 ($\pm5.66$) \\  
8R/40T   & 95.44($\pm4.32$)  & 94.43($\pm4.88$)  & 93.48($\pm4.37$)   & 93.93($\pm5.05$)  & 96.41($\pm3.96$)  & \cellcolor{gray}{96.11$\pm4.56)$}  & 95.24($\pm4.44$)  \\  
8R/50T   & 95.69($\pm3.18$)  & 96.68($\pm2.81$)  & 97.35($\pm4.20$)   & 94.02($\pm2.69$)  & 94.50($\pm4.44$)   & 94.86($\pm3.26$)  & \cellcolor{gray}{96.85$\pm3.40)$}\\\bottomrule 
\end{tabular}}
\end{table*}

\begin{table*}[h!]
  \hspace*{-0.5in}
  \vspace*{-3mm}
\centering
\caption{Training complexity (mean of 20 trials of training, linear \& deterministic env.)}
\scriptsize
\vskip 0.1in
\label{scalability}
\begin{tabular}{cccccccc}
\toprule
\multirow{2}{*}{\textbf{Linear \& Deterministic}} & \multicolumn{7}{c}{\textbf{Testing size : Robot (R) / Task (T)}} \\
 & 2R/20T & 3R/20T & 3R/30T & 5R/30T & 5R/40T & 8R/40T & 8R/50T \\\hline
Performance with full training& 98.31   & 97.50   & 97.80   & 95.35   & 96.99  & 96.11   & 96.85 \\
Training time for 93\%  optimality &19261.2 &61034.0 &99032.7&48675.3&48217.5&45360.0&47244.2\\\bottomrule 
\end{tabular}
\end{table*}

\subsection{Experiment settings}
In the main text, we focus on discrete-time \& discrete-state (DTDS) MRRC problems with deterministic and stochastic task completion times. For CTCS deterministic problems with real-world datasets, see IPMS (Appendix \ref{IPMSappendix}) and mTSP (Appendix \ref{minmaxmtsp}).

\noindent
\textbf{Environment.} Since there is no standard dataset for MRRC problems, we used the  complex maze-like environment generator of \cite{Neller2010} (code provided in Appendix 10). This complex maze mimics the complex road layout of a city and random traffic, inducing nontrivial task completion times. See the leftmost image of Figure \ref{PGM} and the supplementary video. We randomly generated a new maze for every training and testing experiment with randomly chosen initial task/robot locations. To generate the task completion times, Dijkstra's algorithm and dynamic programming were used for deterministic and stochastic environments, respectively. 

In the stochastic environment, a robot makes its intended move with a certain probability. (Cells with a dot: success with 55\%, every other direction with  15\% each. Cells without a dot: 70\% and 10\%, respectively.) A task is considered served when a robot reaches it. We consider two reward rules: linearly decaying rewards $f(age) = \operatorname{max}\{200-age,0\}$ and nonlinearly decaying rewards $f(age)=\lambda^{age}$ with $\lambda$ = 0.99, where $age$ is the task age when served. The initial age of tasks are uniformly distributed in the interval $[0, 100]$.

\noindent
\textbf{Baselines.} For deterministic environments with linear rewards, where the corresponding MRRC can be formulated as a mixed-integer linear program (MILP), we consider the following two baselines:

\begin{itemize}[leftmargin=0.5cm]
\vspace{-0.4cm}
\setlength\itemsep{-0.1em}
    \item \textit{Optimal}: Gurobi \cite{gurobi}, an off-the-self the optimization solver for MILP, was used to solve the problems with 60-min time limit. 
    \item \textit{Ekici et al.}: \cite{Ekici2013}, the most up-to-date heuristic for solving MRRC in the Operations Research community, was used the problems. 
\end{itemize} 
\vspace{-0.4cm}
  For stochastic environments or exponential rewards, to our knowledge, there is no literature addressing MRRC with. Thus, we construct an indirect baseline: 
\begin{itemize}[leftmargin=0.5cm]
\vspace{-0.4cm}
\setlength\itemsep{-0.1em}  
    \item \textit{Sequential Greedy Algorithm (SGA)}:  a general-purpose multi-robot task allocation algorithm called SGA \cite{Choi2009}.
\end{itemize}

The performance measure we used is $\rho =\frac{\textrm{Rewards collected by the proposed method}}{\textrm{Reward collected by the baseline}}$. Thus, the value of $\rho$ greater than 100\% indicates the proposed method collects more reward than the corresponding baseline algorithm. Note that $\rho$ against Optimal is always lower than 100\%.  

Note that we cannot provide other reinforcement learning-based heuristics as additional baselines since, to the best of our knowledge (for the class of NP-hard multi-robot/machine scheduling problems with decaying rewards), this paper is the first to propose a reinforcement learning-based heuristic.

\subsection{Performance test.}  
Performance was tested under four environments: deterministic/linear rewards, deterministic/nonlinear rewards, stochastic/linear rewards, stochastic/nonlinear rewards. See Table \ref{Performance-table}. Our method achieves near-optimality for linear/deterministic rewards with 3\% fewer rewards than \textit{optimal} on average. The standard deviation for $\rho$ is provided in parentheses. For other environments, we see that the \%SGA ratio for linear/deterministic is well maintained. Due to dataset generation's dynamic programming computation complexity, we only consider 8 robots/50 tasks at maximum. We considered larger size problems in IPMS experiments discussed in Appendix A.

\subsection{Transferability test.} 
Table \ref{transfer} provides comprehensive transferability test results. The rows indicate training conditions, while the columns indicate testing conditions. The results in the diagonal cells in red (cells with the same training size and testing size) serve as baselines (direct testing). The results in the off-diagonal show the results for the transferability testing and demonstrate how the algorithms trained with different problem sizes perform well on test problems (zero-shot transfer). We can see that lower-direction transfer tests (trained with larger size problems and tested with smaller size problems) show only a small loss in performance. For upper-direction transfer tests (trained with smaller size problems and tested with larger size problems), the  loss was up to 4 percent.

\subsection{Scalability analysis.} 
For training complexity, we measured the training time required to achieve 93\% optimality considering a deterministic environment with linear rewards. Table 4 shows that training time may not necessarily increase as problem size gets larger, while the performance is fairly maintained. 

MRRC can be formulated as a semi-MDP (SMDP) based multi-robot planning problem (e.g., \cite{Omidshafiei2017}). This problem’s complexity with $R$ robots and $T$ tasks and maximum H time horizon is $O((R!/T!(R-T)!)^H)$. In our proposed method, this complexity is addressed by a combination of two complexities: computational complexity and training complexity. For computational complexity of joint assignment decision at each timestep is $O(|R||T|^3)$. See Appendix \ref{complexity} for details.

\section{Concluding Remarks} In this paper, we addressed the challenge of developing a near-optimal learning-based method for solving NP-hard multi-robot/machine scheduling problems. 
We developed a theory of mean-field inference for scheduling problems and a corresponding  theoretically justified GNN method to precisely infer the Q-function. We addressed the scalability issue of Fitted Q-Iteration methods for multi-robot/machine scheduling problems by providing a polynomial-time algorithm with a provable performance guarantee. Simulation results demonstrate the effectiveness of the our methods. 

\subsection*{Acknowledgement}
Jinkyoo Park was supported by Institute of Information \& communications Technology Planning \& Evaluation (IITP) grant funded by the Korea government(MSIT)(2022-0-01032, Development of Collective Collaboration Intelligence Framework for Internet of Autonomous Things).

\bibliography{references}
\bibliographystyle{icml2021}
\newpage
\section*{Checklist}

The checklist follows the references.  Please
read the checklist guidelines carefully for information on how to answer these
questions.  For each question, change the default \answerTODO{} to \answerYes{},
\answerNo{}, or \answerNA{}.  You are strongly encouraged to include a {\bf
justification to your answer}, either by referencing the appropriate section of
your paper or providing a brief inline description.  For example:
\begin{itemize}
  \item Did you include the license to the code and datasets? \answerYes{See Section~\ref{gen_inst}.}
  \item Did you include the license to the code and datasets? \answerNo{The code and the data are proprietary.}
  \item Did you include the license to the code and datasets? \answerNA{}
\end{itemize}
Please do not modify the questions and only use the provided macros for your
answers.  Note that the Checklist section does not count towards the page
limit.  In your paper, please delete this instructions block and only keep the
Checklist section heading above along with the questions/answers below.

\begin{enumerate}

\item For all authors...
\begin{enumerate}
  \item Do the main claims made in the abstract and introduction accurately reflect the paper's contributions and scope?
    \answerYes{}
  \item Did you describe the limitations of your work?
    \answerYes{}
  \item Did you discuss any potential negative societal impacts of your work?
    \answerNA{}
  \item Have you read the ethics review guidelines and ensured that your paper conforms to them?
    \answerYes{}
\end{enumerate}

\item If you are including theoretical results...
\begin{enumerate}
  \item Did you state the full set of assumptions of all theoretical results?
    \answerYes{}
        \item Did you include complete proofs of all theoretical results?
    \answerYes{}
\end{enumerate}

\item If you ran experiments...
\begin{enumerate}
  \item Did you include the code, data, and instructions needed to reproduce the main experimental results (either in the supplemental material or as a URL)?
    \answerYes{}
  \item Did you specify all the training details (e.g., data splits, hyperparameters, how they were chosen)?
    \answerYes{}
        \item Did you report error bars (e.g., with respect to the random seed after running experiments multiple times)?
    \answerYes{}
        \item Did you include the total amount of compute and the type of resources used (e.g., type of GPUs, internal cluster, or cloud provider)?
    \answerYes{}
\end{enumerate}

\item If you are using existing assets (e.g., code, data, models) or curating/releasing new assets...
\begin{enumerate}
  \item If your work uses existing assets, did you cite the creators?
    \answerNA{}
  \item Did you mention the license of the assets?
    \answerNA{}
  \item Did you include any new assets either in the supplemental material or as a URL?
    \answerNA{}
  \item Did you discuss whether and how consent was obtained from people whose data you're using/curating?
    \answerNA{}
  \item Did you discuss whether the data you are using/curating contains personally identifiable information or offensive content?
    \answerNA{}
\end{enumerate}

\item If you used crowdsourcing or conducted research with human subjects...
\begin{enumerate}
  \item Did you include the full text of instructions given to participants and screenshots, if applicable?
    \answerNA{}
  \item Did you describe any potential participant risks, with links to Institutional Review Board (IRB) approvals, if applicable?
    \answerNA{}
  \item Did you include the estimated hourly wage paid to participants and the total amount spent on participant compensation?
    \answerNA{}
\end{enumerate}

\end{enumerate}

\newpage
\appendix

\onecolumn
\renewcommand{\theequation}{A.\arabic{equation}}
\setcounter{equation}{0}

\section{Identical parallel machine scheduling problem (IPMS) with makespan minimization objective} 
\label{IPMSappendix}
\subsection{Formulation}

IPMS is a problem defined in continuous state/continuous time space. In MRRC, zero processing time of a task was assumed and only the travel times mattered. While there is no travel time concept in IPMS, IPMS has `processing time' and `setup time'. Once service of a task $i$ begins, it requires a deterministic duration of time $\tau_i$ for a machine to complete - we call this the processing time. Machines are all identical, which means processing time of each tasks among machines are all the same. Processing times of each tasks are all different. Before a machine can start processing a task, it is required to first setup for the task. In this paper, we discuss IPMS with ‘sequence-dependent setup times’. In this case, a machine must conduct a setup prior to serving each task. The duration of this setup depends on the current task $i$ and the task $j$ that was previously served on that machine - we call this the setup time. The completion time for each task is thus the sum of the setup time and processing time. Under this setting, we solve the IPMS problem for make-span minimization as discussed in \cite{Kurz2001}. That is, we seek to minimize the total time spent from the start time to the completion of the last task. IPMS problem's sequential decision making problem formulation resembles that of MRRC with continuous-time and continuous-space. That is, every time there is a finished task, we make assignment decision for a free machine. We call this times as `decision epochs' and express them as an ordered set $(t_1, t_2, \dots, t_k, \dots)$. Abusing this notation slightly, we use $(\cdot)_{t_k} = (\cdot)_k$. This problem can be cast as a Markov Decision Problem (MDP) whose state, action, and reward are defined as follows:

\subsection{State}
The state $s_k$ at epoch $k$ is represented as $(g_k, \mathcal{D}_k)$ where a graph $g_k=((\mathcal{M}, \mathcal{T}_k),(\mathcal{E}_{k}^{\mathcal{TT}}, \mathcal{E}_{k}^{\mathcal{MT}}) )$ and associated feature set $\mathcal{D}_k=(\mathcal{D}^{\mathcal{M}}_k,\mathcal{D}^{\mathcal{T}}_k,\mathcal{D}^{\mathcal{TT}}_k,\mathcal{D}^{\mathcal{MT}}_k)$. The elements of graph $g_k$ are defined as: (See Figure 1):
\begin{itemize}[leftmargin=0.5cm]
\vspace{-0.0cm}
\setlength\itemsep{-0.1em}
    \item $\mathcal{M}=\{1,...,M\}$ is the index set of all machines. The index $i$ and $j$ will be used to specifically denote machines in the manuscript.
    \item $\mathcal{T}_k=\{1,...,N\}$ is the index set of all remaining unserved tasks at decision epoch $k$. The index $p$ and $q$ will be used to specifically denote tasks in the manuscript.
    \item $\mathcal{E}_{k}^{\mathcal{TT}}=\{\epsilon_{pq}^{\mathcal{TT}}|p\in\mathcal{T}_k,q\in\mathcal{T}_k\}$ is the set of all directed edges from a task in $\mathcal{T}_k$ to any other task in $\mathcal{T}_k$. Abusing notation slightly, we consider each edge as a random variable. The task-to-task edge $\epsilon_{pq}^{\mathcal{TT}}=1$ indicates the event that a machine that has just completed task $p$ subsequently completes task $q$. We call the probability  $p(\epsilon_{pq}^{\mathcal{TT}}=1)\in[0,1]$ the presence probability of the edge $\epsilon_{pq}^{\mathcal{TT}}$.
    \item $\mathcal{E}_{k}^{\mathcal{MT}}=\{\epsilon_{ip}^{\mathcal{MT}}|i\in\mathcal{M},p\in\mathcal{T}_k\}$ is the set of all directed edges from a machine in $\mathcal{R}$  to any other tasks in $\mathcal{T}_k$. Abusing notation similarly, we say the machine-to-task edge $\epsilon_{ip}^{\mathcal{MT}}=1$ indicates the event that robot $i$ is assigned to the task $p$. This edge is defined deterministically depending on the joint assignment action. If machine $i$ is assigned to task $p$, then $p(\epsilon_{ip}^{\mathcal{MT}})=1$, otherwise 0.
\end{itemize}

The element of feature set $\mathcal{D}_k$ associated with the graph $g_k$ is defined as:
\begin{itemize}[leftmargin=0.5cm]
\vspace{-0.0cm}
\setlength\itemsep{-0.1em}
    \item $\mathcal{D}^{\mathcal{M}}_k=\{d_i^{\mathcal{M}}|i\in \mathcal{M} \}$ is the set of node features for the robot nodes in $\mathcal{M}$ at epoch $k$. In IPMS, $d_i^{\mathcal{R}}$ is defined as the task processing status of robot $i$ at epoch $k$ (epoch index $k$ is omitted).
    \item $\mathcal{D}^{\mathcal{T}}_k=\{d_p^{\mathcal{T}}|p\in \mathcal{T}_k \}$ 	is the set of node features for the task nodes in $\mathcal{T}_k$ at epoch $k$. In IPMS, $d_p^{\mathcal{T}}$ is not used.
    \item $\mathcal{D}^{\mathcal{TT}}_k=\{d_{pq}^{\mathcal{TT}}|p\in\mathcal{T}_k,q\in\mathcal{T}_k\}$ is the set of task-task edge features at epoch $k$. $d_{pq}^{\mathcal{TT}}$ denotes the duration for a machine that has just completed task $p$ to subsequently compete task $q$. 
    We call this duration \textit{task completion time.} In IPMS, a task completion time is sum of processing time and setup time.
    \item $\mathcal{D}^{\mathcal{MT}}_k=\{d_{ip}^{\mathcal{MT}}|i\in\mathcal{M},p\in\mathcal{T}_k\}$ is the set of machine-task edge features at epoch $k$. $d_{ip}^{\mathcal{MT}}$ denotes the task completion time for robot $i$ to reach task $p$. 
\end{itemize}

\subsection{Action}
An action $a_k$, a joint assignment at epoch $k$, is defined as a maximal bipartite matching of the complete bipartite graph $(\mathcal{M}, \mathcal{T}_k,\mathcal{E}_{k}^{\mathcal{MT}})$ composed of the machine nodes $\mathcal{M}$, the remaining task nodes $\mathcal{T}_k$, and the fully connected edges between them $\mathcal{E}_{k}^{\mathcal{MT}}$. That is, given the current state $s_k=(g_k,\mathcal{D}_k)$, $a_k$ is a subset of  $\mathcal{E}_{k}^{\mathcal{MT}}$ satisfying (1) no two machines can be assigned to the same tasks, and (ii) a machine may only remain without assignment when the number of machines exceeds the number of remaining tasks. If $\epsilon_{ip}^{\mathcal{RT}}\in a_k$, it means that machine $i$ is assigned with task $p$ at epoch $k$. For example, Figure 1 shows the case where $a_k=(\epsilon_{1,1}^{\mathcal{RT}},\epsilon_{2,3}^{\mathcal{RT}})$. 
(Note, we equivalently may say this as $\epsilon_{1,1}^{\mathcal{RT}}=\epsilon_{2,3}^{\mathcal{RT}}=1$ and $0$ otherwise.) In IPMS, only free machines are newly assigned.

\subsection{State transition}
When the joint assignment $a_k$ is executed given the current state $s_k=(g_k,\mathcal{D}_k)$  , the next state $s_{k+1}=(g_{k+1},\mathcal{D}_{k+1})$ will be determined. The details differ depending on the problem. 
\\
\textbf{Graph update} when the decision epoch corresponds to the point when task $p$ is completed, the corresponding task node will be removed in the updated task nodes as
$\mathcal{T}_{k+1}=\mathcal{T}_{k}/\{p\}$, and the task-task edges and machine-task edges, $\mathcal{E}_{k+1}^{\mathcal{TT}}$ and $\mathcal{E}_{k+1}^{\mathcal{RT}}$, will be accordingly updated.\\ 
\textbf{Feature update} At decision epoch $k+1$,  $\mathcal{D}_{k+1}=(\mathcal{D}^{\mathcal{R}}_{k+1},\mathcal{D}^{\mathcal{T}}_{k+1},\mathcal{D}^{\mathcal{TT}}_{k+1},\mathcal{D}^{\mathcal{RT}}_{k+1})$ is determined. 
The machine-task edge features $\mathcal{D}^{\mathcal{MT}}_{k+1}$ will be updated according to $\mathcal{D}^{\mathcal{M}}_{k+1}$ as well. How these features are updated is determined by the problem specifications (environment).

\subsection{Reward and objective}
Define an assignment policy $\phi$ as a function that maps a state $s_k$ to action $a_k$. Denote $T\left(s_{k}, a_{k}, s_{k+1}\right)$ as the time difference between epoch $k$ and $k+1$ according to $(s_{k}, a_{k}, s_{k+1})$. Given $s_0$ initial state, an IPMS problem with makespan minimization objective can be expressed as a problem of finding an optimal assignment policy $\phi^*$ such that 
\begin{align*}
    \phi^{*}=\underset{\phi}{\operatorname{argmin}}\; \mathbb{E}_{\phi  }\left[\sum_{k=0}^{\infty} T\left(s_{k}, a_{k}, s_{k+1}\right) | s_{0}\right].
\end{align*}

\begin{table}
\small
\centering
\caption{IPMS test results for makespan minimization with deterministic task completion time (our algorithm / best Google OR tool result)}
\label{ipms}
\begin{tabular}{c|c|cccc}
\hline
                   \multicolumn{2}{c|}{\textbf{Makespan}}     & \multicolumn{4}{c}{\textbf{\# Machines}}  \\\cline{3-6}          
                   \multicolumn{2}{c|}{\textbf{minimization}}    & 3       & 5       & 7        & 10       \\\hline 
\multirow{3}{*}{\textbf{\# Tasks}} & 50  & 106.7\% & 117.0\% & 119.8\%  & 116.7\%  \\ \cline{2-6}
                          & 75  & 105.2\% & 109.6\% & 113.9\%  & 111.3\% \\ \cline{2-6}
                          & 100 & 100.7\% & 111.0\% & 109.1\% & 109.0\%  
\end{tabular}
\end{table}

\subsection{Experiments} For IPMS, we test it with continuous time, continuous state environment. While there have been many learning-based methods proposed for (single) robot scheduling problems, to the best our knowledge our method is the first learning method to claim scalable performance among machine-scheduling problems. Hence, in this case, we focus on showing comparable performance for large problems, instead of attempting to show the superiority of our method compared with heuristics specifically designed for IPMS (actually no heuristic was specifically designed to solve our exact problem (makespan minimization, sequence-dependent setup with no restriction on setup times)) 

For each task, processing times is determined using uniform [16, 64]. For every (task $i$, task $j$) ordered pair, a unique setup time is determined using uniform [0, 32]. As illustrated in Appendix \ref{IPMSappendix}, we want to minimize make-span. As a benchmark for IPMS, we use Google OR-Tools library \cite{Google2012}. This library provides metaheuristics such as Greedy Descent, Guided Local Search, Simulated Annealing, Tabu Search. We compare our algorithm's result with the heuristic with the best result for each experiment. We consider cases with $3, 5, 7, 10$ machines and $50, 75, 100$ jobs.

The results are provided in Appendix Table \ref{ipms}. Makespan obtained by our method divided by the makespan obtained in the baseline is provided. Although our method has limitations in problems with a small number of tasks, it shows comparable performance to a large number of tasks and shows its value as the first learning-based machine scheduling method that achieves scalable performance.

\section{minimax multiple traveling salesman problem (minimax mTSP)}\label{minmaxmtsp}
\subsection{Formulation}

Minimax multiple traveling salesman problem (minimax mTSP) is almost the same problem as IPMS problem defined in Appendix \ref{IPMSappendix}. As in IPMS, we seek to minimize the total time spent from the start time to the completion of the last task. (Remark: minimax mTSP problem's original objective is to minimize the longest tour of all the salesmen. As the traveling speed of all the salesmen are identical, this objective is equivalent to makespan minimization objective of IPMS problem.) One difference in minimax mTSP from IPMS problem is that it's task completion time is just traveling time from a task to another task. Another difference is the existence of a `depot' in minimax mTSP problem; every salesman start from the depot at time $0$ and must return to the depot in the end.  

IPMS problem's sequential decision making problem formulation is the same as that of IPMS. That is, every time there is a finished task, we make assignment decision for a free machine. We call this times as `decision epochs' and express them as an ordered set $(t_1, t_2, \dots, t_k, \dots)$. Abusing this notation slightly, we use $(\cdot)_{t_k} = (\cdot)_k$. This problem can be cast as a Markov Decision Problem (MDP) whose state, action, and reward are almost exactly the same as that of IPMS except that we call a machines of IPMS a salesman.

\subsection{Estimating state-action value function}
In minimax mTSP, we don't delete the tasks that were previously served. 

As in MRRC, two hierarchical layers of \textit{random structure2vec} is used to infer the $Q$-function. Recall that in MRRC, the only input of the first random structure2vec was each task's robot assignment information. In minimax mTSP, we add three more information as input: each task's distance from the depot, each task's coordinate, and whether the task has been served by then.

The second structure2vec is the same as that of MRRC.

Given the output vectors of second structure2vec, we separately sum the vectors for the tasks that are not yet served and vectors for the tasks that are yet served. Given two separately summed output vectors, we concatenate the two resulting vectors and estimate the $Q$-function. 

\subsection{Experiments} 
Achieving multiple-TSP performance similar to the single-TSP result of famous \cite{Dai2017} (90\%-92\% optimal) qualifies itself a separate conference or journal paper. In this paper, we briefly introduce the reader the capability of our proposed method to solve minimax mTSP in a similar level of performance to \cite{Dai2017}. In this experiment, we encoded relational information among nodes using TrXL-I (Parisoto et al., 2020) before computing presence probability. After encoding relational information among nodes, we used this information as node features for our model and computed presence probability with the edge features.

\textbf{Dataset.} We used the standard minmax mTSP dataset and state-of-art optimal solution baselines for them provided in \cite{minmaxTSPlib}. Each problem in the dataset is named after a city, where task locations in each city are originated from real world locations. For example, Berlin52 problem means that the task locations are originated from 52 real locations of the city of Berlin in Germany. 
\cite{minmaxTSPlib} provides the state-of-art solution (found by integer linear programming model solved by CPLEX or reported by others) that minimizes the longest length tour. 

Table \ref{table:mTSPtable} shows proposed algorithm's performance compared with the result provided in \cite{minmaxTSPlib}. 

\begin{table}
\centering
\caption{Test results on \cite{minmaxTSPlib}}
\label{table:mTSPtable}
\begin{tabular}{c|c|ccc}
\textbf{Problem name}              &  \# agents           & CPLEX  & OR-Tool & Ours   \\
\toprule
\multirow{4}{*}{\textbf{eli51}}    & \textbf{M=2} & 222.7  & 243.4   & 233.4  \\
                                   & \textbf{M=3} & 159.6  & 170.1   & 171.9  \\
                                   & \textbf{M=5} & 124.0  & 127.5   & 131.7  \\
                                   & \textbf{M=7} & 112.1  & 112.1   & 114.8  \\
                                   \toprule
\multirow{4}{*}{\textbf{berlin52}} & \textbf{M=2} & 4110.2 & 4665.5  & 4313.9 \\
                                   & \textbf{M=3} & 3244.4 & 3311.3  & 3243.5 \\
               
                                  & \textbf{M=5} & 2441.4 & 2482.6  & 2638.5 \\
                                   & \textbf{M=7} & 2440.9 & 2440.9  & 2474.1 \\
                                   \toprule
\multirow{4}{*}{\textbf{eli76}}    & \textbf{M=2} &  280.9  &  318.0  & 298.8 \\
                                   & \textbf{M=3} &  197.3  &  212.4  &  215.6 \\
                                   & \textbf{M=5} &  150.3  &  143.4  &  158.4 \\
                                   & \textbf{M=7} &  139.62 &  128.3  &  140.8 \\
                                   \toprule
\multirow{4}{*}{\textbf{rat99}}    & \textbf{M=2} & 733.8 &  762.2  & 728.7 \\
                                   & \textbf{M=3} & 592.6 &  552.1  & 587.2 \\
                                   & \textbf{M=5} & 502.9 &  473.7  & 469.3 \\
                                   & \textbf{M=7} & 473.1 &  442.5  & 443.9 \\
                                   \toprule
                                   & \textbf{Average ratio} & 1      & 1.0180  & 1.0268
\end{tabular}
\end{table}

Table \ref{table:mTSPtable} compares the outcome of our method and the state-of-art solution provided by \cite{minmaxTSPlib} and Google OR-Tool\cite{Google2012}. The state-of-art solution. Our proposed method's solution achieves in average $2.68$\% sub-optimality, which is the first to achieve a comparable result to Google OR-Tool ($1.80\%$). We can see that the optimal solution has within 5\% less cost than our proposed method's solution, which is much better than the sub-optimality of \cite{Dai2017} for single-traveling salesman problem.

\section{Proof of Theorem 1.} \label{Thm1pf}
We first define necessary definitions for our proof.
Given a random PGM $\{\mathcal{G}_\mathcal{X}, \mathcal{P}\}$, a PGM is chosen among $\mathcal{G}_\mathcal{X}$, the set of all possible PGMs on $\mathcal{X}$. The set of semi-cliques is denoted as $\mathfrak{C}_{\mathcal{X}}$. As discussed in the main text, if we are given $\mathcal{P}$ then we can easily calculate the presence probability $p_m$ of semi-clique $\mathcal{D}_{m}$ as $p_m = \sum_{G\in\mathcal{G}_\mathcal{X}}\mathcal{P}(G)1_{\mathcal{D}_{m}\in G}$. 

For each semi-clique $\mathcal{D}^{i}$ in $\mathfrak{C}_{\mathcal{X}}$, define a binary random variable $V^i$: $\mathcal{F}\mapsto \{0, 1\}$ with value 0 for the factorization that does not include semi-clique $\mathcal{D}^{i}$ and value 1 for the factorization that include semi-clique $\mathcal{D}^{i}$. Let $V$ be a random vector $V=\left(V^{1}, V^{2}, \ldots, V^{|\mathfrak{C}_{\mathcal{X}}|}\right)$. Then we can express $P(X_1,\ldots,X_n|V) \propto \prod_{i=1}^{|\mathfrak{C}_{\mathcal{X}}|}\left[\phi^{i}\left(\mathcal{D}^{i}\right)\right]^{V^{i}}$. We denote $ \left[\phi^{i}\left(\mathcal{D}^{i}\right)\right]^{V^{i}}$ as $\psi(\mathcal{D}^{i})$.

Now we prove Theorem 1.

In mean-field inference, we want to find a distribution $Q\left(X_{1}, \ldots, X_{n} \right)=\prod_{i=1}^n Q_{i}(X_{i})$ such that the cross-entropy between it and a target distribution is minimized. Following the notation in \cite{Koller}, the mean field inference problem can written as the following optimization problem. 
\begin{align*}
\min_{Q} \quad& \mathbb{D}\left(\prod_{i} Q_{i}\left|P\left(X_{1}, \ldots, X_{n} | V\right)\right)\right)\\
\textrm{s.t.} \quad& \sum_{x_{i}} Q_{i}\left(x_{i}\right)=1 \quad \forall i\\
\end{align*}
Here $\mathbb{D}\left(\prod_{i} Q_{i} \;\middle|\; P\left(X_{1}, \ldots, X_{n} | V\right)\right)$ can be expressed as 
 $\mathbb{D}\left(\prod_{i} Q_{i} \;\middle|\; P\left(X_{1}, \ldots, X_{n} | V\right)\right)=\mathbb{E}_{Q}\left[\ln \left(\prod_{i} Q_{i}\right)\right]-\mathbb{E}_{Q}\left[\ln \left(P\left(X_{1}, \ldots, X_{n} | V\right)\right)\right]$. \\
Note that 
\begin{align*}
    \mathbb{E}_{Q}&\left[\ln \left(P\left(X_{1}, \ldots, X_{n} | V\right)\right)\right]=\mathbb{E}_{Q}\left[\ln \left(\frac{1}{z} \Pi_{i=1}^{|\mathfrak{C}_{\mathcal{X}}|} \psi^{i}\left(\mathcal{D}^{i}, V\right)\right)\right]\\
     &=\mathbb{E}_{Q}\left[\ln \left(\frac{1}{z} \prod_{i=1}^{|\mathfrak{C}_{\mathcal{X}}|} \psi^{i}\left(\mathcal{D}^{i}, V\right)\right)\right] \\ &=\mathbb{E}_{Q}\left[\sum_{i=1}^{|\mathfrak{C}_{\mathcal{X}}|} V^{i} \ln \left(\phi^{i}\left(\mathcal{D}^{i}\right)\right)\right]-\mathbb{E}_{Q}[\ln (Z)]
     \\ &=\sum_{i=1}^{|\mathfrak{C}_{\mathcal{X}}|}\mathbb{E}_{Q}\left[V^{i} \ln \left(\phi^{i}\left(\mathcal{D}^{i}\right)\right)\right]-\mathbb{E}_{Q}[\ln (Z)] 
     \\& =\sum_{i=1}^{|\mathfrak{C}_{\mathcal{X}}|}\mathbb{E}_{V^{i}}\left[\mathbb{E}_{Q}\left[V^{i} \ln  \left(\phi^{i}\left(\mathcal{D}^{i}\right)\right) | V^{i}\right]\right]-\mathbb{E}_{Q}[\ln (Z)]  \\
     & =\sum_{i=1}^{|\mathfrak{C}_{\mathcal{X}}|} P\left(V^{i}=1\right)\left[\mathbb{E}_{Q}\left[\ln  \left(\phi^{i}\left(\mathcal{D}^{i}\right)\right)\right]\right]-\mathbb{E}_{Q}[\ln (Z)] \\
     & =\sum_{i=1}^{|\mathfrak{C}_{\mathcal{X}}|}p_{i}\left[\mathbb{E}_{Q}\left[\ln \left(\phi^{i}\left(\mathcal{D}^{i}\right)\right)\right]\right]-\mathbb{E}_{Q}[\ln (Z)].
\end{align*}

Hence, the above optimization problem can be written as 

\begin{equation}\label{eq:2}
\begin{aligned}
\max_{Q} \quad& \mathbb{E}_{Q}\left[\sum_{i=1}^{|\mathfrak{C}_{\mathcal{X}}|}p_{i}\ln \left(\phi^{i}\left(\mathcal{D}^{i}\right)\right)\right]+\mathbb{E}_{Q}\sum_{i=1}^{n}\left(\ln  Q_{i}\right) \\
\textrm{s.t.} \quad& \sum_{x_{i}} Q_{i}\left(x_{i}\right)=1 \quad \forall i\\
\end{aligned}
\end{equation}

In \cite{Koller}, the fixed point equation is derived by solving an analogous equation to (\ref{eq:2}) without the presence of the $p_i$. Theorem 1 follows by proceeding as in \cite{Koller} with straightforward accounting for $p_i$. 

\section{Hilbert space embedding of distributions.}\label{embedding}
We start from the motivation of Hilbert space embedding of distributions, with a particular focus on mean-field inference application. As discussed in section 3, mean-field inference methods try to search over the space of distributions, looking for the best surrogate distribution. While exact optimal solution search is a open, difficult optimization problem, at least we know that the optimal distribution must satisfy a fixed point equation we saw in section 3 \cite{Koller}. While this is only a necessary condition and does not bring us an optimal solution, in practice distributions that satisfies such condition works as a nice approximate solution. Nevertheless, finding a distribution that satisfies the fixed equation of distribution involves an intractable equation of integrals. 

A Hilbert space embedding of distributions transforms this kind of optimization problems over distributions into optimization problems over a vector space. Suppose that a random vector $X$ is associated with a  joint distribution $F$. Then for a function $\phi$ on the range of $X$, we can define a mapping towards a Hilbert space defined as $\mu_{X}:=\mathbb{E}_{X}[\phi(X)]=\int \phi(x) dF(x)$. This kind of operation was first introduced in 
\cite{smola2007hilbert}. According to \cite{sriperumbudur2008injective}, there exist some $\phi$ that makes this operation an injective operation. Therefore, when we map the entire fixed point iteration on the distribution space to the Hilbert space, we don't lose any mathematical structure.

\section{Proof of Lemma 1.} \label{Lem1pf}
Since we assume semi-cliques are only between two random variables, we can denote $\mathfrak{C}_{\mathcal{X}}$ = $\{\mathcal{D}^{ij}\}$ and presence probabilities as $\{p_{ij}\}$ where $i, j$ are node indexes. Denote the set of nodes as $\mathcal{V}$.

From here, we follow the approach of \cite{Dai2016} and assume that the joint distribution of random variables can be written as 
\begin{align*}
    p\left(\left\{H_{k}\right\},\left\{X_{k}\right\}\right) \propto \prod_{k \in \mathcal{V}} \psi^{i}\left(H_{k} | X_{k}\right) \prod_{k, i \in \mathcal{V}} \psi^{i}\left(H_{k} | H_{i}\right).
\end{align*}

Expanding the fixed-point equation for the mean field inference from Theorem 1, we obtain: 
\begin{align*}
    &Q_{k} \left(h_{k}\right) = \\ &\frac{1}{Z_{k}} \exp \left\{\sum_{\psi^{i} : H_{k} \in \mathcal{D}^{i}} \mathbb{E}_{\left(\mathcal{D}^{i}-\left\{H_{k}\right\}\right) \sim Q}\left[\ln \psi^{i}\left(H_{k}=h_{k} | \mathcal{D}^{i}\right)\right]\right\}\\
    &=\frac{1}{Z_{k}} \exp \{ln \phi\left(H_{k}=h_{k} | x_{k}\right)+ \\&\sum_{i \in \mathcal{V}} \int_{\mathcal{H}} p_{k i} Q_{i}\left(h_{i}\right) \ln \phi\left(H_{k}=h_{k} | H_{i} \right) d h_{i} \}.
\end{align*}
This fixed-point equation for $Q_{k}\left(h_{k}\right)$ is a function of $\left\{Q_{j}\left(h_{j}\right)\right\}_{j \neq k}$ such that
\begin{align*}
    Q_{k}\left(h_{k}\right)=f\left(h_{k}, x_{k},\left\{p_{k j} Q_{j}\left(h_{j}\right)\right\}_{j \neq k}\right).
\end{align*}
As in \cite{Dai2016}, this equation can be expressed as a Hilbert space embedding of the form
\begin{align*}
    \tilde{\mu}_{k}=\tilde{\mathcal{T}} \circ\left(x_{k},\left\{p_{k j} \tilde{\mu}_{j}\right\}_{j \neq i}\right),
\end{align*}
where $\tilde{\mu}_{k}$ indicates a vector that encodes $Q_{k}\left(h_{k}\right) .$ In this paper, we use the nonlinear mapping $\tilde{\mathcal{T}}$ (based on a neural network form ) suggested in \cite{Dai2016}:
\begin{align*}
    \tilde{\mu_{k}}=\sigma\left(W_{1} x_{k}+W_{2} \sum_{j \neq k} p_{k j} \tilde{\mu}_{j}\right)
\end{align*}

\newpage
\section{Presence probability inference method used for MRRC} \label{pinference}

In this experiment, we encoded relational information among nodes using TrXL-I style Multi-Head Self-Attention structure \cite{parisotto2020stabilizing} to compute presence probability. To compute a better presence probability (which is closer to the optimal solution), it is essential to consider relational information among nodes. Node features are stacked and then passed to Multi-Head Self-Attention, which encodes the relational information between nodes. After encoding relational information among nodes, we used these relational node features as the original node features for our GNN model and computed presence probability with these relational node features and the edge features. The rest of the part for estimating the Q-function is identical to the MRRC problem.

\section{Complete algorithm of section \ref{Qinference} with task completion time as a random variable} \label{randomsample}

We combine random sampling and inference procedure suggested in section and Figure 2. Denote the set of task with a robot assigned to it as $\mathcal{T}^A$. Denote a task in $\mathcal{T}^A$ as $t_i$ and the robot assigned to $t_i$ as $r_{t_i}$. The corresponding edge in $\mathcal{E}^{RT}$ for this assignment is $\epsilon_{r_{t_i}t_i}$. The key idea is to use samples of $\epsilon_{r_{t_i}t_i}$ to generate $N$ number of sampled $Q(s, a)$ value and average them to get the estimate of $E(Q(s, a))$. First, for $l = 1\dots N$ we conduct the following procedure. For each task $t_i$ in $\mathcal{T}^A$, we sample one data $e_{r_{t_i}t_i}^l$. Using those samples and $\{p_{ij}\}$, we follow the whole procedure illustrated in section \ref{embedding} to get $Q(s, a)^l$. Second, we get the average of $\{Q(s, a)^l\}_{l=1}^{l=N}$ to get the estimate of $E(Q(s, a))$,  $\frac{1}{N}\sum_{l=1}^{l=N}Q(s, a)^l$.

The complete algorithm of section \ref{embedding} with task completion time as a random variable is given as below.

1   $\;\; age_i=$ age of node $i$\\
\hspace{0.1cm}2   $\;\;$\textit{The set of nodes for assigned tasks} $\equiv \mathcal{T}_{A}$\\
\hspace{0.1cm}$\;$ 3   $\;\;$\textit{Initialize} $\{\tilde{\mu}_{i}^{(0)}\},\{\gamma_{i}^{(0)}\}$\\
\hspace{0.1cm}$\;$ 4   $\;\;$ for $l=1$ to $N$:\\
$\;$ 5   $\;\;\quad $ for $t_i\in \mathcal{T}$:\\
5   $\;\;\quad \quad$ if $t_i\in \mathcal{T^A}$ do:\\
6   $\;\;\quad \quad \quad$ sample $e_{r_{t_i}t_i}^l$ from $\epsilon_{r_{t_i}t_i}$ \\
7   $\;\;\quad \quad \quad$ $x_i = e_{r_{t_i}t_i}^l$\\ 
\hspace{0.2cm} 9   $\;\;\quad \quad$ else: $x_i = 0$\\
10  $\;\;\quad$ for $t=1$ to $T_1$ do\\
11  $\;\;\quad \quad$ for $i\in \mathcal{V}$ do\\
12  $\;\;\quad \quad\quad \; l_i = \sum_{j \in \mathcal{V}} p_{j i} \tilde{\mu}_{j}^{(t-1)}$\\
13  $\;\;\quad \quad\quad \; \tilde{\mu}_{i}^{(t)}=r e l u\left(W_{3} l_{i}+W_{4} x_{i}\right)$\\
14  $\;\;\quad\; \widetilde{\mu}_{l}=\text { Concatenate }\left(\tilde{\mu}_{i}^{(T_{1})}, age_i\right)$ \\
15  $\;\;\quad$ for $t=1$ to $T_{2}$ do\\
16  $\;\;\quad\quad$ for $i \in \mathcal{V}$ do\\
17  $\;\;\quad\quad \quad \; l_{i}=\sum_{j \in \mathcal{V}} p_{j i} \gamma_{j}^{(t-1)}$\\
18  $\;\;\quad\quad \quad \; \gamma_{j}^{(t)}=relu\left(W_{5} l_{i}+W_{6} \tilde{\mu}_{i}\right)$\\
19  $\;\;\quad\; Q_{l}=W_{7} \sum_{i \in \mathcal{V}} \gamma_{i}^{(T)}$\\
20  $\;\; Q_{a v g}=\frac{1}{N} \sum_{l=1}^{N} Q_{l}$

\newpage
\section{Proof of Lemma 2} \label{Lem2pf}
\textbf{Statement:} \textit{Denote result of OTAP using true Q-functions $\{Q^{(n)}\}$ as $\mathcal{M}^{(N)}$ $=$ $\{m^{(1)} \dots m^{(N)}\}$. If $Q$-function approximation method has order transferability, then $\mathcal{M}^{(N)}$ = $\mathcal{M}_\theta^{(N)}$ holds.
}\\
\textbf{Proof.} Recall that we say Q-function approximation method has order transferability if $\operatorname{argmax}_{a_{t_k}} Q^n(s_{t_k}, a_{t_k})$ = $\operatorname{argmax}_{a_{t_k}} Q^n_{\theta}(s_{t_k}, a_{t_k})$. We prove by induction. 
\\
Base case: For $n=0$,   $\mathcal{M}^{(0)}=\phi=  \mathcal{M}_\theta^{(0)}$. 
\\For $n>0$, suppose that $ \mathcal{M}^{(n)}=\mathcal{M}_\theta^{(n)}$ holds, i.e. $m^{(j)} = m_\theta^{(j)}$ for $1\le j\le n$. Then according to $n+1^{th}$ step OTEP operation, \\
$m^{(n+1)}=\operatorname{argmax}_{m} Q^{n+1}\left(s_{t_{k}}, \mathcal{M}^{(n)}\cup \{m\}\right)$$\\
=$$\operatorname{argmax}_{m} Q_\theta^{n+1}\left(s_{t_{k}}, \mathcal{M}^{(n)}\cup \{m\}\right) (\because $ Order transferability assumption$)$\\
= $\operatorname{argmax}_{m} Q_\theta^{n+1}\left(s_{t_{k}}, \mathcal{M}_\theta^{(n)}\cup \{m\}\right) (\because $ induction argument$)$\\
= $m_\theta^{(n+1)}$. 
\\Therefore, $\mathcal{M}^{(n+1)}=\mathcal{M}^{(n)} \cup\{m^{(n+1)}\} = \mathcal{M}_\theta^{(n)} \cup\{m_\theta^{(n+1)}\} = \mathcal{M}_\theta^{(n+1)}$.

\section{Statement and Proof of Lemma 3.}\label{lemma3} Denote the space of all possible policies as $\Pi$. For $\pi \in \Pi$, let the vector $d_t^{\pi}$ denote the distribution of states at arbitrary time $t$ assuming that we have been following policy $\pi$ from time $0$. We call a policy $\mu \in \Pi$ is exploratory with respect to $\Pi$ if $\exists C<\infty$ such that $\forall \pi \in \Pi, \frac{d_t^{\pi}(s)}{\mu(s)} \leq C$ holds $\forall s \in \mathcal{S}$ and $\forall t\ge0$. The assumption that such exploratory policy $\mu$ exists is called \textit{Concentrability} assumption. Recall that in Auction-fitted Q-iteration (AFQI) we want to find $\theta$ that empirically minimizes $E_{(s_k, a_k, r_k, s_{k+1})\sim D }$ $\left[Q_{\theta}\left(s_{k}, a_{k}\right)-\left[r\left(s_{k}, a_{k}\right)+\gamma Q_{\theta}\left(s_{k+1}, \pi_{Q_{\theta}}\left(s_{k+1}\right)\right)\right]\right]$ where $D$ is a dataset. Denote the set of possible $Q$-functions as  $\mathcal{F}\subset \mathbb{R}^{\mathcal{S}\times\mathcal{A}}$. In our case, $\mathcal{F} = \{Q_\theta\}$. If we denote an operation $\mathcal{T}$ on the $\mathcal{F}$ such that $\mathcal{T}f(s, a) =: E_{(s, a, r, s^{\prime})\sim D }$ $\left[\left[r+\gamma f\left(s, \pi_f\left(s^{\prime}\right)\right)\right]\right],$ our problem can be restated as finding $ f^*=\operatorname{argmin}_{f\in \mathbb{R}^{\mathcal{S}\times\mathcal{A}}} (f - \mathcal{T}f)$. We say that this problem is \textit{realizable} if $f^*\in \mathcal{F}$. We say that $\mathcal{F}$ is \textit{closed under Bellman update} if $\forall f \in \mathcal{F}, \mathcal{T} f \in \mathcal{F}$. For details of above assumptions, see \cite{agarwal2019reinforcement}. 

We now formally state Lemma 3. 

\textit{\underline{Lemma 3 \cite{kang2021approximate}}}. Under the assumption that concentrability, realizability and closure under Bellman update holds, the policy we achieve by AFQI is assured to have performance at least $1-1/e$ compared with the optimal policy.

\section{Decentralized algorithm}\label{decentralized} In this section, we show that we can modify the auction procedure in OTAP at each timestep as a special case of \cite{Choi2009}’s sequential greedy algorithm for solving MRTA problem. This enables us to conclude, without further discussion, that 1) assignment consensus is guaranteed among robots even under frequent communication packet loss and 2) centralized algorithm’s performance bound is inherited. The Decentralized algorithm is almost the same as the centralized version. Bolded sentences indicate what is different in decentralized version of suggested algorithm.
    
\underline{Initial message choice phase.}
In the $n^{th}$ bidding phase, initially \textit{all robots know}  $\mathcal{M}_\theta^{(n-1)}$, the ordered set of $n-1$ robot-task edges in $\mathcal{E}_{t_{k}}^{R T}$ determined by the previous $n-1$ iterations. An unassigned robot $i$ ignores all others unassigned and calculates $Q_\theta^n(s_{t_{k}},\mathcal{M}_\theta^{(n-1)} \cup \{ \epsilon_{ip}^{RT} \} )$ for each unassigned task $p$ as if those $k$ robots (robot $i$ together with all robots assigned tasks in the previous $n-1$ iterations) only exist in the future and will serve all remaining tasks. (Here, $\epsilon_{ip}^{RT}\in\mathcal{E}_{t_k}^{RT}$ is the edge corresponding to assigning robot $i$ to task $p$ at decision epoch $t_k$.) If task $\ell$ has the highest value, robot $i$ chooses $\{\epsilon_{i \ell}^{RT},Q_\theta^n(s_t, \mathcal{M}_\theta^{(n-1)}\cup \{ \epsilon_{i \ell}^{RT} \} )\}$ as the initial message to be sent to others. (Note that, since the number of ignored robots varies at each iteration, transferability of Q-function inference is crucial)

\underline{Consensus phase.} In the $n^{th}$ consensus phase, robot $i$ keeps sending message to neighbouring robots within its one-hop communication range. At first, agent $i$ keeps sending its initial message chosen in above phase to neighbors. Then every time a robot $i$ receives a message $\{\epsilon_{j m}^{RT},Q_\theta^n(s_t, \mathcal{M}_\theta^{(n-1)}\cup \{ \epsilon_{j m}^{RT} \} )\}$ of robot $j$, it compares $Q_\theta^n(s_t, \mathcal{M}_\theta^{(n-1)}\cup \{ \epsilon_{jm}^{RT} \} )$ with $Q_\theta^n(s_t, \mathcal{M}_\theta^{(n-1)}\cup \{ \epsilon_{i \ell}^{RT} \} )$. If the former is larger, robot $i$ sets $\{\epsilon_{j m}^{RT},Q_\theta^n(s_t, \mathcal{M}_\theta^{(n-1)}\cup \{ \epsilon_{j m}^{RT} \} )\}$ as the new message to be sent to others. Agent $i$ keeps sending its new message to neighbors until it hears all robot's initial messages. Denote the message of $i$ right after it hears all robot's initial message as $m_\theta^{(n)}$ (Note that the initial message of $i$ includes robot $i$'s assignment information). Then agent $i$ updates the assignment set as $\mathcal{M}_\theta^{(n)}=\mathcal{M}_\theta^{(n-1)} \cup m_\theta^{(n)}$.

\section{Proof of Theorem 2} \label{Thm2pf}
\textbf{Statement:} \textit{Denote $N=\max \left(|\mathcal{R}|,\left|T_{t}\right|\right)$. \\ Suppose that $Q$-function approximation method has order transferability. Denote $\mathcal{M}_{\theta}^{(N)}$ as the result of OTAP using $\left\{Q_{\theta}^{n}\right\}$ and $\mathcal{M}^{*}$ as $\operatorname{argmax}_{a_{t_{k}}} Q\left(s_{t_{k}}, a_{t_{k}}\right)$.
If 1) the marginal value of adding one robot is positive, i.e. $Q^{|\mathcal{M}|+1}(s_{t_k}, \mathcal{M} \cup\{m\})-Q^{|\mathcal{M}|}(s_{t_k}, \mathcal{M}) \ge 0$ for all $\mathcal{M} \subset \mathcal{E}_{t}^{R T}$ and 2) the marginal value of adding one robot diminishes as the robot number increases, i.e., $Q^{|\mathcal{M}|+1}(s_{t_k}, \mathcal{M} \cup\{m\})-Q^{|\mathcal{M}|}(s_{t_k}, \mathcal{M} ) $ 
$\le$ 
$Q^{|\mathcal{N}|+1}(s_{t_k}, \mathcal{N} \cup\{m\})-Q^{|\mathcal{N}|}(s_{t_k}, \mathcal{N}) $ for $\mathcal{N} \subset \mathcal{M}\subset \mathcal{E}_{t}^{R T}$, for all $m \in \mathcal{E}_{t}^{R T}$, then the result of OTAP is at least better than $1-1/e$ of optimal assignment, i.e., $Q_{\theta}^{N}(s_{t_{k}}, \mathcal{M}_{\theta}^{(N)}) \ge Q^{|\mathcal{M}^{*}|}\left(s_{t_{k}}, \mathcal{M}^{*}\right)\left(1-1/e\right).$}\\ 

\textbf{Proof.} 
From the assumption 1) that the marginal value of adding one robot is nonnegative, without loss of generality, we can consider $\mathcal{M}^{*}$ with $|\mathcal{M}^{*}| = N$ in the further proof procedure. Denote  $\mathcal{M}^{*}$$=$$\{m^{(1)*}, m^{(2)*},\dots, m^{(n)*}\}$ and denote $\mathcal{M}_{\theta}^{(N)}=\{m_{\theta}^{(1)}, m_{\theta}^{(2)},\dots, m_{\theta}^{(N)}\}$.
\\For notation simplicity, define  
$\Delta(m\mid \mathcal{M})=:$  $Q^{|\mathcal{M} \cup\{m\}|}(s_t, \mathcal{M} \cup\{m\})-Q^{|\mathcal{M}|}(s_t, \mathcal{M}).$

Then the optimal value $OPT$ = $Q^{N}(s_{t_{k}}, \mathcal{M}^{*})
\le 
Q^{|\mathcal{M}_\theta^{(n)}\cup\mathcal{M}^{*}|}(s_{t_{k}}, \mathcal{M}_\theta^{(n)}\cup\mathcal{M}^{*})$
\\$=Q^{n}(s_{t_{k}}, \mathcal{M}_\theta^{(n)})+ \sum_{j=1}^{N}\Delta(m^{(j)*}\mid \mathcal{M}_\theta^{(n)}\cup\{m^{(1)*}, \cdots, m^{(j-1)*}\} )$
\\ $\le Q^{n}(s_{t_{k}}, \mathcal{M}_\theta^{(n)})+ \sum_{j=1}^{N}\Delta(m^{(j)*}\mid \mathcal{M}_\theta^{(n)}) (\because$ condition 2 - decreasing marginal value condition$)$ 
\\
$\le Q^{n}(s_{t_{k}}, \mathcal{M}_\theta^{(n)})+ \sum_{j=1}^{N}\Delta(m_{\theta}^{(n+1)}\mid \mathcal{M}_\theta^{(n)}) 
\\
(\because$ OTAP chooses $m_{\theta}^{(n+1)}=\operatorname{argmax}_{m} Q_{\theta}^{n+1}\left(s_{t}, \mathcal{M}_{\theta}^{(n)} \cup\left\{m\right\}\right)$ and 
\\
$\operatorname{argmax}_{m} Q_{\theta}^{n+1}\left(s_{t}, \mathcal{M}_{\theta}^{(n)} \cup\left\{m\right\}\right) = \operatorname{argmax}_{m} Q^{n}\left(s_{t}, \mathcal{M}_{\theta}^{(n)} \cup\left\{m\right\}\right)$ from \textit{Lemma 2})
\\
$=Q^{n}(s_{t_{k}}, \mathcal{M}_\theta^{(n)})+ N\Delta(m_{\theta}^{(n+1)}\mid \mathcal{M}_\theta^{(n)})$. \\ 
Therefore, $\Delta(m_{\theta}^{(n+1)}\mid \mathcal{M}_\theta^{((n))}) \ge \frac{1}{N}(OPT - Q^{n}(s_{t_{k}}, \mathcal{M}_\theta^{(n)})$.\\
Note that $OPT - Q^{n}(s_{t_{k}}, \mathcal{M}_\theta^{(n)})$ denotes current iteration $(=n^{th})$ outcome $\mathcal{M}_{\theta}^{(n)}$'s size of sub-optimality compared to $OPT$. Denote $OPT - Q^{n}(s_{t_{k}}, \mathcal{M}_\theta^{(n)})=:\beta_n$. Then since $Q^{0}(s_{t_{k}}, \phi)=0$, $\beta_0 = OPT$. Therefore, we have $\Delta(m_{\theta}^{(n+1)}\mid \mathcal{M}_\theta^{((n))}) \ge \frac{1}{N}\beta_n$. \\
Also, note that $\Delta(m_{\theta}^{(n+1)}\mid \mathcal{M}_\theta^{(n)}) = Q^{n+1}(s_t, \mathcal{M}_\theta^{(n)} \cup\{m_{\theta}^{(n+1)}\})-Q^{n}(s_t, \mathcal{M}_\theta^{(n)})$
\\
$=Q^{n+1}(s_t, \mathcal{M}_\theta^{(n+1)})-Q^{n}(s_t, \mathcal{M}_\theta^{(n)}) = (OPT-Q^{n}(s_t, \mathcal{M}_\theta^{(n)})-(OPT-Q^{n+1}(s_t, \mathcal{M}_\theta^{(n+1)}))$
\\
$=\beta_n-\beta_{n+1}$.\\
Therefore, $\beta_n-\beta_{n+1} \ge \frac{1}{N}\beta_n$, i.e., $\beta_{n+1} \leq \beta_{n}\left(1-\frac{1}{N}\right)$.\\
This implies $OPT-Q^{N}(s_{t_{k}}, \mathcal{M}_{\theta}^{(N)})=\beta_{N} \le \beta_0(1-\frac{1}{N})^N = OPT (1-\frac{1}{N})^N$ and thus we get \\
$Q^{N}(s_{t_{k}}, \mathcal{M}_{\theta}^{(N)}) = OPT(1-(1-\frac{1}{N})^N) \sim OPT(1-\frac{1}{e})$ as ${N\rightarrow \infty}$.

\newpage
\section{Scalability analysis}\label{complexity}
\textbf{Computational complexity}. MRRC can be formulated as a semi-MDP (SMDP) based multi-robot planning problem (e.g., \cite{Omidshafiei2017}). This problem’s complexity with $R$ robots and $T$ tasks and maximum H time horizon is $O((R!/T!(R-T)!)^H)$. For example, \cite{Omidshafiei2017} state that a problem with only 13 task completion times (`TMA nodes’ in their language) possessed a policy space with cardinality $5.622*10^{17}$. In our proposed method, this complexity is addressed by a combination of two complexities: computational complexity and training complexity. For computational complexity of joint assignment decision at each timestep, it is $O(|R||T|^3)=O((1)\times(2)\times(3)\times(4)+(5))$ where $(1)-(5)$ are as follows. 
\begin{itemize}
\itemsep0em
    \item[(1)] \# of Q-function computation required in one time-step = $O(|R||T|)$: Shown in section 4.2
    \item[(2)] \# of mean-field inference in one Q-function computation = 2 (constant): Two embedding steps (Distance embedding, Value embedding) each needs one mean-field inference procedure
    \item[(3)] \# of structure2vec propagation operation in one mean-field inference= $O(|T|^2)$: There is one structure2vec operation from a task to another task and therefore the total number of operations is $|T|\times(|T|-1)$.
    \item[(4)] \# of neural net computation for each structure2vec propagation operation=C (constant):  This is only dependent on the hyperparameter size of neural network and does not increase as number of robots or tasks.
    \item[(5)] \# of neural net computation for inference of random PGM=$O(|T|^{2})$ As an offline stage, we infer the semi-clique presence probability for every possible directed edge, i.e. from a task to another task using algorithm introduced in Appendix $\ref{pinference}$. This algorithm complexity is $O(|T|\times(|T|-1)) = O(|T|^2)$.
\end{itemize}

\end{document}